\documentclass[10pt,a4paper]{article}
\usepackage[T1]{fontenc}
\usepackage[utf8]{inputenc}
\usepackage{amssymb}
\usepackage[standard-baselineskips]{cmbright}
\usepackage[sfdefault,semibold,type1]{plex-sans}
\usepackage[tracking=true]{microtype}
\SetTracking{encoding={OT1,OML,OMS,OMX,U}}{0}
\SetTracking{encoding=T1,family={IBMPlexSans-TLF,plxSans-TLF}}{-15}
\DeclareMathSizes{10.3}{11.124}{7.7868}{5.562}
\DeclareMathSizes{7}{7.56}{5.4}{5.4}
\DeclareMathSizes{8}{8.64}{6.48}{5.4}
\DeclareMathSizes{9}{9.72}{6.48}{5.4}
\DeclareMathSizes{10}{10.8}{7.56}{5.4}
\DeclareMathSizes{12}{12.96}{8.64}{6.48}
\AtBeginDocument{\lsstyle}
\usepackage[a4paper,top=17mm,bottom=19mm,left=21mm,right=21mm,
            headheight=19pt,headsep=5mm,footskip=10mm]{geometry}

\usepackage{graphicx}
\graphicspath{{figs/}}
\usepackage{amsmath}
\usepackage{siunitx}
\usepackage{booktabs,tabularx,array,multirow}
\usepackage{threeparttable}
\usepackage{makecell}
\usepackage{subcaption}
\usepackage{placeins}
\usepackage{float}
\usepackage{enumitem}
\usepackage{titlesec}
\usepackage{fancyhdr}
\usepackage[most]{tcolorbox}
\usepackage{needspace}
\usepackage{cite}
\usepackage{xcolor}
\usepackage{xurl}
\usepackage{hyperref}
\definecolor{SignalInk}{HTML}{2F3135}
\definecolor{SignalHeading}{HTML}{1C236C}
\definecolor{SignalBar}{HTML}{335DD1}
\definecolor{SignalTeal}{HTML}{335DD1}
\definecolor{SignalDarkTeal}{HTML}{1C236C}
\definecolor{SignalMist}{HTML}{F7F9FF}
\definecolor{SignalLine}{HTML}{DCEAFE}
\definecolor{SignalMuted}{HTML}{1C236C}
\definecolor{SignalCaption}{HTML}{5B637A}

\newcommand{\newtext}[1]{#1}

\newcommand{\PaperTitle}{Expert-level detection of low-voltage intervals in EEG as a robust biomarker of neonatal encephalopathy}
\newcommand{\PaperShortTitle}{Detecting LVIs in Neonatal EEG}
\newcommand{\PaperType}{Research article}

\newcommand{\PaperVersion}{Preprint\enspace\textbullet\enspace October 2026}
\newcommand{\PaperKeywords}{neonatal EEG, inter-burst interval, low-voltage interval, encephalopathy, deep learning, clinical biomarker}

\newcommand{\PaperAuthorsDisplay}{John M. O'Toole\textsuperscript{1}, Sean R. Mathieson\textsuperscript{1,2}, Aurel Luca\textsuperscript{1}, Soraia Ventura\textsuperscript{1,2}, Sean Griffin\textsuperscript{1}, Geraldine B. Boylan\textsuperscript{1,2,3}, and Robert Hogan\textsuperscript{1}}

\newcommand{\PaperAffiliationsDisplay}{%
  \textsuperscript{1}CergenX Ltd, Cork, Ireland\par
  \textsuperscript{2}INFANT Research Centre, University College Cork, Ireland\par
  \textsuperscript{3}Department of Paediatrics and Child Health, University College Cork, Ireland}

\newcommand{\PaperCorrespondence}{jotoole@cergenx.com, rhogan@cergenx.com}

\newcommand{\AbstractBackground}{Discontinuity in neonatal EEG is a critical marker of brain dysfunction but can be difficult to quantify through visual review. An automated, artificial intelligence (AI)-based approach has the potential to quantify discontinuity in an objective and repeatable manner. Inter-burst intervals (IBIs) are a key characteristic of discontinuity, yet their definition does not easily translate to AI techniques.}

\newcommand{\AbstractObjectives}{To introduce low-voltage intervals (LVIs) as a context-independent alternative to IBIs, defined exclusively by amplitude and duration. To develop an AI detection method for LVIs and determine their clinical utility as a marker of brain injury.}

\newcommand{\AbstractMethods}{We developed a deep-learning model (ConvNeXt) to detect LVIs and compared its performance with existing quantitative methods. We then tested the clinical relevance of the detected LVIs across four applications: 1) grading background EEG abnormalities on an open benchmark dataset, 2) classifying clinical HIE grades and comparing with a baseline normal EEG dataset, 3) seizure prediction, and 4) association with seizure burden.}

\newcommand{\AbstractResults}{For training and validation of the detection methods, \qty{654}{\hour} of per-channel EEG from 123 neonates were reviewed, resulting in \num{51162} LVI annotations. On held-out validation, the ConvNeXt method had an area under the receiver operating characteristic curve (AUC) of 0.996 and a Matthews correlation coefficient of 0.926. The ConvNeXt model agreed more with the individual EEG readers than the EEG readers agreed with each other ($\Delta \kappa = 0.025$, 95\% CI: 0.007 to 0.045, $p<0.001$), thus attaining expert-level equivalence. On a combined dataset of 444 neonates, we found a strong association with percentage LVI across EEG grades (Spearman's correlation coefficient $\rho=0.801$), a moderate association with HIE grades ($\rho=0.384$), a weaker association with seizure burden ($\rho=0.244$), and a strong seizure prediction capacity (AUC $=0.872$).}

\newcommand{\AbstractConclusion}{AI methods to detect LVIs achieve expert-level equivalence, whereas quantitative features do not. Automated quantification of LVIs provides a robust clinical biomarker of neonatal encephalopathy with clear associations with background EEG activity, clinical HIE severity, and seizure risk.}

\hypersetup{
  colorlinks=true,
  linkcolor=SignalDarkTeal,
  citecolor=SignalDarkTeal,
  urlcolor=SignalDarkTeal,
  pdftitle={Expert-level detection of low-voltage intervals in EEG as a robust biomarker of neonatal encephalopathy},
  pdfauthor={John M. O'Toole et al.},
  pdfsubject={Low-voltage interval detection in neonatal EEG},
  pdfkeywords={neonatal EEG, inter-burst interval, low-voltage interval, encephalopathy, deep learning, clinical biomarker},
  pdfdisplaydoctitle=true
}

\color{SignalInk}
\setlist{leftmargin=*,topsep=3pt,itemsep=2pt}

\renewcommand{\headrulewidth}{0.35pt}
\renewcommand{\headrule}{\hbox to\headwidth{\color{SignalLine}\leaders\hrule height \headrulewidth\hfill}}

\titleformat{\section}
  {\needspace{4\baselineskip}\Large\bfseries\color{SignalHeading}}
  {\color{SignalHeading}\makebox[9mm][l]{\thesection}}
  {0pt}{}
  [\vspace{3pt}{\color{SignalLine}\titlerule}]
\titleformat{name=\section,numberless}
  {\needspace{4\baselineskip}\Large\bfseries\color{SignalHeading}}
  {}{0pt}{}
  [\vspace{3pt}{\color{SignalLine}\titlerule}]
\titlespacing*{\section}{0pt}{2.0em}{0.75em}

\titleformat{\subsection}
  {\needspace{3\baselineskip}\large\bfseries\color{SignalHeading}}
  {\color{SignalHeading}\thesubsection}
  {0.65em}{}
\titleformat{name=\subsection,numberless}
  {\needspace{3\baselineskip}\large\bfseries\color{SignalHeading}}
  {}{0pt}{}
\titlespacing*{\subsection}{0pt}{1.45em}{0.35em}

\titleformat{\subsubsection}
  {\needspace{2.5\baselineskip}\normalsize\bfseries\color{SignalHeading}}
  {\color{SignalHeading}\thesubsubsection}
  {0.65em}{}
\titleformat{name=\subsubsection,numberless}
  {\needspace{2.5\baselineskip}\normalsize\bfseries\color{SignalHeading}}
  {}{0pt}{}
\titlespacing*{\subsubsection}{0pt}{1.2em}{0.25em}

\AddToHook{env/figure/begin}{%
  \ifnum\value{figure}=0
    \setlength{\intextsep}{26pt plus 4pt minus 3pt}%
    \captionsetup{skip=13pt}%
  \fi
}

\newcommand{\PaperReferencesHeading}{}
\newenvironment{PaperDemographicsTable}{\begin{table}[H]}{\end{table}}
\AddToHook{env/table/begin}{\let\scriptsize\small}
\AddToHook{env/tablenotes/begin}{\let\scriptsize\footnotesize\footnotesize}
\newcommand{\PaperDatasetLink}{%
  \href{https://zenodo.org/records/7477575}{Zenodo record 7477575}}

\newcommand{\PaperPartDivider}[2]{%
  \par\Needspace{10\baselineskip}%
  \par\vspace{1.1em}%
  \noindent
  \begin{tcolorbox}[
    colback=SignalBar,colframe=SignalBar,boxrule=0pt,arc=0pt,
    left=4mm,right=4mm,top=2.2mm,bottom=2.2mm]
    \color{white}\bfseries\MakeUppercase{#1}\hfill
    \normalfont\color{SignalMist}#2
  \end{tcolorbox}
  \par\nobreak\vspace{0.2em}\nobreak%
}

\newcommand{\abstractrow}[2]{%
  \textcolor{SignalTeal}{\bfseries\footnotesize\MakeUppercase{#1}} &
  \small #2\tabularnewline[4pt]
}

\newcommand{\PaperTitleBadge}[1]{%
  \tcbox[on line,colback=white!9!SignalBar,colframe=white!15!SignalBar,
    boxrule=0.3pt,arc=1mm,boxsep=0pt,
    left=2mm,right=2mm,top=1.4mm,bottom=1.4mm]{%
    \fontsize{7}{9}\selectfont\mdseries\color{white}\strut{}#1}%
}

\newcommand{\makepaperfront}{%
  \thispagestyle{empty}%
  \begingroup
  \vspace*{-10mm}%
  \begin{tcolorbox}[
    enhanced,width=\dimexpr\textwidth+30mm\relax,
    enlarge left by=-15mm,enlarge right by=-15mm,
    colback=SignalBar,colframe=SignalBar,boxrule=0pt,arc=1.5mm,
    boxsep=0pt,left=15mm,right=15mm,top=8mm,bottom=8mm,
    before skip=0pt,after skip=5mm]
    \color{white}%
    {\fontsize{7.5}{10}\selectfont
      \PaperTitleBadge{\PaperVersion}\par}
    \vspace{4mm}
    {\raggedright\hyphenpenalty=10000\fontsize{26}{28}\selectfont
      \bfseries\color{white}\PaperTitle\par}
    \vspace{5mm}
    {\fontsize{10.3}{13.4}\selectfont\bfseries\color{white}\PaperAuthorsDisplay\par}
    \vspace{3mm}
    \noindent
    \begin{minipage}[b]{\dimexpr0.65\linewidth-6mm\relax}
      \fontsize{7.5}{10}\selectfont\raggedright
      \setlength{\parskip}{0pt}\lineskiplimit=-10pt\relax
      \color{white}\PaperAffiliationsDisplay\par
    \end{minipage}\hspace{6mm}%
    \begin{minipage}[b]{0.35\linewidth}
      \fontsize{7.5}{10}\selectfont\raggedright
      \setlength{\parskip}{0pt}
      \color{white}\textbf{Correspondence}\par\PaperCorrespondence\par
    \end{minipage}
  \end{tcolorbox}
  \endgroup

  \begin{tcolorbox}[
    colback=SignalMist,colframe=SignalMist,boxrule=0pt,arc=0pt,
    left=5mm,right=5mm,top=4mm,bottom=3mm]
    {\small\bfseries\color{SignalHeading}ABSTRACT}\par
    \vspace{2mm}{\color{SignalLine}\hrule}\vspace{3mm}
    \begin{tabularx}{\linewidth}{@{}p{21mm}@{\hspace{3mm}}X@{}}
      \abstractrow{Background}{\AbstractBackground}
      \abstractrow{Objectives}{\AbstractObjectives}
      \abstractrow{Methods}{\AbstractMethods}
      \abstractrow{Results}{\AbstractResults}
      \abstractrow{Conclusion}{\AbstractConclusion}
    \end{tabularx}
  \end{tcolorbox}
  \vfill
  {\footnotesize\color{SignalMuted}\textbf{Keywords}\quad \PaperKeywords\par}
  \vspace{2mm}{\color{SignalTeal}\rule{\textwidth}{0.7pt}}
  \clearpage
}

\begin{document}
\fontsize{10.3}{13.4}\selectfont
\makepaperfront

\section{Introduction}
\label{sec:intro}

Neonatal encephalopathy is a clinical syndrome of disturbed neurological function caused by an acute or chronic insult to the brain \cite{Schendel2014}.
It is a leading cause of neonatal death and lifelong neurological impairment \cite{Lawn2005,Tan2025}.
Early detection will enable timely interventions, such as therapeutic hypothermia for hypoxic-ischaemic encephalopathy (HIE).
The electroencephalogram (EEG) is the clinical gold standard for monitoring brain function and can be used to assess and track evolving injuries and neurological recovery.
Yet EEG requires specialist readers to visually interpret the complex waveforms and patterns.
Expertise in interpretation of neonatal EEG is scarce and rarely available 24/7.
Artificial intelligence (AI) has the potential to assist with EEG review as it excels at pattern matching tasks such as identifying specific waveforms within the signal.
AI can also scale to match massive volumes of data generated by long-duration monitoring across multiple neonates.

The application of AI to neonatal EEG has primarily centred on seizure detection.
Automated classification of background EEG abnormalities, however, has received more attention in recent years \cite{Moghadam2022,OtextquotesingleSullivan2023,Twomey2025} with a particular focus on the clinical application \cite{Montazeri2024,Tuiskula2024,Kota2024,Lagace2024,GonzalezTamez2025,Cornet2025}.
Stratifying the different grades of severity of background activity has the potential to provide a useful metric of overall brain health that is more consistent than the episodic nature of seizures.
Unfortunately, there is no generally accepted grading scheme for neonatal EEG \cite{Fanning2025}.
Amplitude and discontinuity are the only common factors across all grading schemes \cite{Walsh2011,Fanning2025}.
Inter-burst intervals (IBIs), which are periods of low-voltage or suppressed activity between bursts, are the defining characteristic of discontinuity.
Automated detection of IBIs is challenging because the clinical context dictates interpretation: voltage thresholds change depending on vigilance states or background activity patterns.
For example, IBIs are defined as activity \qty{<25}{\micro\volt} for tracé discontinu and within \qtyrange{25}{50}{\micro\volt} for tracé alternant \cite{Tsuchida2013}.
Thus far, existing IBI detection methods have largely ignored this context, training and testing on data from a single vigilance state or background pattern only \cite{Mirzaei2017,Raurale2019,Raurale2020,Lundy2023}.

In this study, we focus on low-voltage activity itself, without requiring a burst--interburst pattern.
We use the term low-voltage interval (LVI) to describe a period of EEG activity meeting prespecified low-amplitude and duration criteria in an individual channel.
An LVI does not require identifiable preceding or subsequent bursts and may be confined to a single channel or occur concurrently across multiple channels.
Its extent across channels is therefore characterised separately from its identification.
An LVI may coincide with an IBI when the relevant criteria are met, but the two terms are not interchangeable.
We develop machine-learning (ML) methods that are trained and validated on a large and diverse dataset encompassing different types of EEG vigilance states and background patterns, comparing with existing state-of-the-art methods.
We then establish automated LVI detection as a clinically relevant biomarker of neonatal encephalopathy across four applications: 1) grading EEG background on an open benchmark, 2) classifying clinical HIE severity against a healthy baseline cohort, 3) predicting seizures, and 4) correlating LVI features with total seizure burden.

\section{Methods}
\label{sec:methods}

\begin{figure}[!ht]
  \centering
  \includegraphics[width=0.98\textwidth]{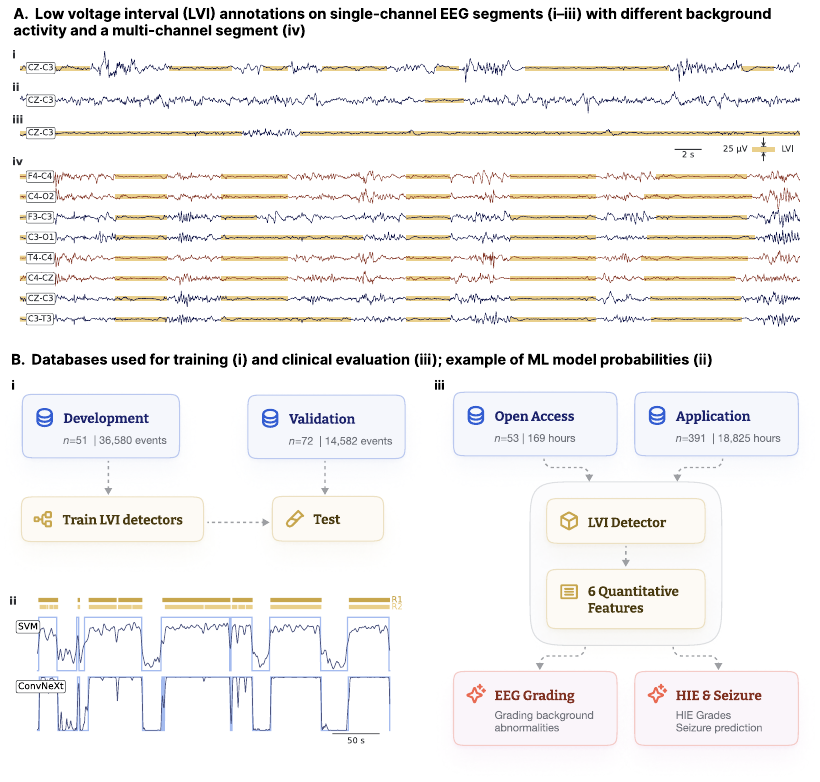}
  \caption{Examples of low-voltage interval (LVI) annotations (A) and datasets used for detector development and clinical evaluation (B).
Panels A(i--iii) show LVI annotations within different patterns of background activity in a single EEG channel (CZ-C3).
Panel A(iv) shows an example of time-synchronised LVIs.
All EEG segments are 60 s and use the same amplitude scale.
Panels B(i) and B(iii) show flowcharts of the datasets used for detector development and clinical evaluation, respectively.
Panel B(ii) shows examples of the probability and binary outputs from two machine-learning models: SVM, a support vector machine, and ConvNeXt, a convolutional deep-learning network.
Two readers (R1 and R2) annotated the LVIs in this example.
}\label{fig:overview_lvis}
\end{figure}

We define LVIs as intervals of activity below \qty{25}{\micro\volt} for 2 or more seconds on an individual channel.
(Voltages are reported as peak-to-peak amplitudes.)  These events are not tied to specific vigilance states or limited to specific background patterns of activity.
Unlike IBIs, they do not necessarily require a preceding or subsequent bursts.
We choose to call these events LVIs rather than IBIs for the following reasons:
\begin{itemize}
\item In tracé alternant, IBIs are defined as low-voltage periods between bursts with an amplitude of \qtyrange{25}{50}{\micro\volt} \cite{Tsuchida2013}; therefore these events would not be included in our definition of LVIs.
\item In tracé discontinu, IBIs are defined as \qty{<25}{\micro\volt} and therefore will overlap completely with our LVI definition; however, the same is true for suppressed activity (\qty{<5}{\micro\volt}) in burst suppression and electrocerebral inactivity (\qty{<2}{\micro\volt}) in an isoelectric EEG.
\item Continuous or nearly-continuous background activity may contain occasional periods of low-voltage activity (\qty{<25}{\micro\volt} for 2 or more seconds) which meet our definition for an LVI.
\end{itemize}
Although our definition of LVIs is based solely on amplitude and independent of EEG state or background pattern, the decision to annotate an event lies with the EEG reader, who uses their experience and judgement to make this determination.
Fig.~\ref{fig:overview_lvis}A presents examples of LVIs in tracé discontinu (Fig.~\ref{fig:overview_lvis}A-i), nearly-continuous EEG (Fig.~\ref{fig:overview_lvis}A-ii), and burst suppression (Fig.~\ref{fig:overview_lvis}A-iii).
Fig.~\ref{fig:overview_lvis}A-iv shows another property of LVIs---they can be both synchronous and asynchronous across channels.

We developed automated methods to detect these LVIs and then tested the association of LVIs to
different markers of neonatal encephalopathy in a large EEG dataset.
For clarity, the following sections are divided into two parts: Part A describes the LVI detection methods, while Part B presents the application of LVI detection to determine clinical relevance.

\PaperPartDivider{Part A}{LVI detection}

\subsection{EEG data and LVI annotation}
\label{sec:eeg_data}

EEG used for the development of the automated detection methods was recorded in the Cork University Maternity Hospital, Ireland, as part of different retrospective research studies.
Informed parental consent was obtained for EEG recordings and all studies received approval from the
Clinical Research Ethics Committee of the Cork Teaching Hospitals.
EEG data, which consisted of long-duration, multi-channel recordings, were fully anonymised before transfer for analysis in this study.
Electrode placement followed a neonatal version of the international 10--20 recording system, across central (C3/Cz/C4), frontal (F3/F4 or Fp1/Fp2), occipital (O1/O2) or parietal (P3/P4), and temporal (T3/T4) locations \cite{Hogan2025}.

Different datasets were used for the development and validation of the LVI detection methods, with no overlap in neonates' EEG.
Table~\ref{tab:train_test_set} presents the summarised demographic and clinical information for these datasets.
The development set was used for training and testing the ML models.
One-hour epochs were pruned from the long-duration continuous EEG recordings from 51 neonates.
Only epochs with minimal artefacts were selected.
For the validation set, \qty{20}{\minute} epochs from 72 neonates were selected, again with minimal artefact.
Both sets included neonates with a confirmed diagnosis of HIE and healthy neonates with EEG recorded in the post-natal ward \cite{Korotchikova2016,Hogan2025}.

A clinical physiologist with expertise in reviewing neonatal EEG annotated LVIs on a per-channel basis for the development set.
All bipolar channels available in the recording were annotated.
Two other clinical physiologists independently annotated LVI events on the validation dataset.
To maximise cohort size within practical annotation constraints, only two channels (C4--Cz and Cz--C3) were annotated for validation.

\begin{PaperDemographicsTable}
  \centering
  \scriptsize
  \caption{Demographic information on the EEG data sets used to develop and validate the
    low-voltage interval (LVI) detectors.}
  \label{tab:train_test_set}
  \begin{threeparttable}
    \begin{tabular}{lrr}
      \toprule
                                             & development set           & validation set             \\
                                             & (n=51)                    & (n=72)                     \\ 
      \midrule
      Gestational Age (weeks)                & 40+2 (39+2 -- 40+5)       & 39+6 (39+0 -- 40+5)        \\
      Birth Weight (kg)                      & 3.5 (3.33 -- 3.85) [n=26] & 3.64 (3.37 -- 3.83) [n=23] \\
      Sex, Female                            & 18 (35.3\%)               & 28 (38.9\%)                \\
      HIE                                    & 26 (51.0\%)               & 55 (76.3\%)                \\
      HIE: Mild / Moderate / Severe          & 9 / 13 / 4                & 26 / 18 / 11               \\
      Normal cohort                          & 25 (49.0\%)               & 17 (23.6\%)                \\
      \midrule
      Total EEG duration (hours)             & 80.3                      & 33.4                       \\
      Total EEG per-channel duration  & 587.05                    & 66.7
      \\
       (hours) &                     &                        \\      
      \bottomrule
    \end{tabular}
    \begin{tablenotes}
      \scriptsize
      \item HIE: hypoxic ischaemic encephalopathy
    \end{tablenotes}
  \end{threeparttable}
\end{PaperDemographicsTable}

\subsection{Detection Methods}
\label{sec:detection_methods}

To develop an automated LVI detection method, we started by exploring existing methods for detecting IBIs.
Several of these methods were originally developed to detect bursts in preterm EEG, but can also be used to detect IBIs, as the absence of a burst in preterm discontinuous activity inherently implies an IBI.
A number of these burst detectors have previously been adapted to detect IBIs in term EEG \cite{Mirzaei2017,Raurale2019,Raurale2020}.
A common approach for IBI detection is to use quantitative features to process the EEG and then apply a threshold to determine IBI or burst.
Another approach is to use ML methods, a successful strategy in almost all other EEG detection applications.
A possible reason why ML methods have not been used as extensively in this application is the assumption that the task is not complex enough and a simple feature should suffice.
We test this hypothesis by comparing feature-based approaches with an existing ML method for detecting IBIs \cite{O’Toole2017} and propose a new approach using a convolutional neural network recently developed for neonatal EEG seizure detection \cite{Hogan2025}.

\subsubsection{Quantitative Features}

Four features used to detect bursts or IBIs are included as baseline methods \cite{Jennekens2011,O’Toole2014,Koolen2014,Matic2016,Mirzaei2017}.
These features are defined in more detail in Appendix~\ref{sec-app:quant-feat-ibi}.
Two features are amplitude based: a peak-to-peak measure, defined in (\ref{eq:5}), which calculates the ratio of activity outside the \qty{\pm 12.5}{\micro\volt} range; and the envelope of the signal, defined in (\ref{eq:3}).
The remaining features are frequency-weighted energy features: the envelope-derivative operator, defined in (\ref{eq:2}), which is an improved variant of the common non-linear energy operator \cite{Palmu2010,Palmu2010a,Matic2016,Mirzaei2017}; and line-length, defined in (\ref{eq:4}).

A \qty{2}{\second} median filter is applied to smooth all features except for line-length, which inherently incorporates moving-average smoothing.
The threshold for each feature is determined by balancing sensitivity and specificity.

\subsubsection{Machine Learning}

Here we implement an ML model that has shown its versatility in detecting bursts or IBIs in both term and preterm EEG \cite{O’Toole2017,Raurale2019,Raurale2020}.
The method uses eight quantitative EEG features, including the envelope-derivative operator defined in (\ref{eq:2}), and combines these features in a linear support vector machine (SVM).
The features extract information on instantaneous amplitude, instantaneous frequency, and spectral shape.
Incorporating multiple features in a trained ML model typically performs better than a single feature alone \cite{O’Toole2017}.

This SVM model was trained on our development dataset (Section~\ref{sec:eeg_data}).
The unbounded SVM output is converted to a probability through a sigmoid function and this continuous value is then convolved with a \qty{2.25}{\second} rectangular window.
The binary mask, generated by applying a 0.5 threshold to the probability function, is post-processed to drop positive (LVI) or negative (non-LVI) events \qty{<1}{\second}.
This post-processing procedure was developed through empirical evaluation during $k$-fold cross-validation.

We also develop a neural network model to detect LVIs.
Specifically, we use a ConvNeXt architecture \cite{Liu2022} that was recently adapted for neonatal EEG seizure detection \cite{Hogan2025}.
We use the Nano version of the model, a small model with 38.7k trainable parameters, as, unlike for the seizure detection model, we found no further gains by scaling model size.
The model was trained on \qty{2}{\second} segments of EEG with a time-shift of \qty{0.25}{\second}.
Data augmentation during training included signal inversion (flip), cutout \cite{DeVries2017}, and a time-warping method \cite{Hogan2025}.
The network's parameters were trained using the AdamW optimiser and a learning-rate scheduler with 4 distinct phases.
The phases---warm-up, freeze at maximum, cool-down, and freeze at minimum---were similar to those for the XL seizure model but compressed to fit within the fixed 30 training epochs.
Post-processing was limited to smoothing with a 1 s rectangular window and removing positive and negative events \qty{<1}{\second}.

For both the ML methods and quantitative features, the EEG is band-pass filtered within the range 0.3 to 30 Hz and downsampled to 64 Hz.
Both approaches are trained and tested on the development set using 10-fold cross-validation.
For the features, the threshold is set during training.
Finally, the algorithms are trained on the entire development set before testing on the validation set.

\subsubsection{Detection performance}
\label{sec:detect-perf}

We used a battery of sample-based metrics to evaluate and compare detection performance
\cite{Hogan2025,Kljajic2025}.
Performance is evaluated both by concatenating the output across neonates and on a per-neonate
basis.
Where two readers reviewed the EEG, a consensus annotation was used by discarding the samples where there was disagreement between readers.

On the validation set, we assessed whether the methods passed an expert-level equivalence test.
This test determines whether agreement changes significantly when one reader is replaced by the detection method \cite{Tapani2022,Hogan2025,Kljajic2025}.
For this, we generated a Cohen's $\kappa$ from the readers' annotations ($\kappa_{\text{readers}}$) and compared it to the $\kappa$ generated by swapping each reader's annotation, in turn, with the detection method's annotation ($\kappa_{\text{detector}}$).
We then created a $\Delta \kappa = \kappa_{\text{readers}} - \kappa_{\text{detector}}$ distribution by bootstrapping across neonates.
To pass the test, the lower margin of the $\Delta \kappa$ distribution, the 2.5th percentile, needed to be above a preset threshold.
However, instead of setting an arbitrary threshold to allow for an accepted decrease in performance for a non-inferiority test, we set this threshold to the margin of error from the $\kappa_{\text{readers}}$ distribution.
This represents a lower bound on expected inter-rater agreement from the readers and therefore reframes the non-inferiority test as an expert-level equivalence test.

\PaperPartDivider{Part B}{Application of LVI detection}

To validate the automated detection methods (\emph{LVI detectors}) and to demonstrate the potential clinical importance of LVIs, we apply the best-performing detection method to an array of clinical EEG data.
We test how well detected LVIs distinguish (a) EEG grades of background abnormalities, (b) clinical
HIE grades, and (c) neonates who later develop seizures, and (d) association with seizure burden.

\subsection{Application EEG datasets}

For the first clinical evaluation, we use an open-access dataset of EEG with grades of the severity of abnormal background activity.
This dataset contains 169 \qty{1}{\hour} epochs of multi-channel EEG from 53 neonates.
Further details on this dataset can be found in \cite{OToole2023}.

For the remaining three applications, we used an anonymised dataset of EEG sourced from previous research studies at the Cork University Maternity Hospital (Ireland) and the University College London Hospital (UK).
Ethical approval was granted for the original studies, with written informed consent obtained from parents or guardians.
Retrospective analysis of the anonymised data was approved by the Clinical Research Ethics Committee of the Cork Teaching Hospitals.
This dataset consisted of \qty{18824}{\hour} of multi-channel recordings for 391 neonates.
See Table~\ref{tab:evaluation_set} in the Appendix for demographic details.

There was no overlap between the EEG data from the development set, used to train the LVI detectors,
and the EEG used in these clinical evaluation experiments.
Fig.~\ref{fig:overview_lvis}B gives an overview of how the datasets were used to develop and evaluate the LVI detectors.

\subsection{Application Experiments}
\label{sec:app_lvi}

To summarise the LVI binary mask across the recording, we used six quantitative features to determine the main characteristics of the temporal mask.
These were the maximum and median duration of an event, the percentage of LVI activity, the number of events, and complexity and periodicity of the events.
For the maximum duration feature, we used the 95th percentile to avoid maximum outliers \cite{Raurale2019,Raurale2020}.
We calculated complexity using the Lempel--Ziv algorithm, which quantifies the number of unique patterns in the binary mask \cite{Ziv1978}.
This measured the level of temporal order or randomness of the LVI events.
Periodicity was measured through the autocorrelation function of the binary mask.
This feature computed the mean absolute correlation across a 10 min lag window and evaluated the extent to which similar LVI patterns recurred over time compared with random noise.

These six LVI features were then applied to the different classification tasks in this evaluation set.
First, we compared the features against four EEG grades that classify background activity as normal/mild, moderate, major, or severe abnormalities.
This open-access dataset contains an EEG grade for each \qty{1}{\hour} epoch of EEG \cite{OToole2023}.
Second, we compared the features against the mild, moderate, and severe grades of HIE.
These grades were assigned from the modified Sarnat score for HIE; if the neonate had
more than one Sarnat score in the first 24 hours, the most severe score was selected \cite{Pavel2022}.
We extracted up to four \qty{1}{\hour} epochs from the continuous EEG at random time points across the long-duration EEG recordings.
This random sampling across time also enforced a \qty{>12}{\hour} gap between epochs.
For comparison, we also included a set of normal EEGs \cite{Korotchikova2016}.
Third, for the seizure prediction task, we extracted the first hour of the EEG recording and excluded neonates with seizure onset \qty{<4}{\hour} from the start of the EEG recording to provide at least a \qty{3}{\hour} gap between the end of the epoch and the start of seizures.
And fourth, to examine the association with seizure burden, we used the entire EEG recording.
For this, we calculated LVI features across a \qty{1}{\hour} epoch with a 75\% overlap and then summarised (median) across all epochs per neonate.
Seizure burden was classified into 4 categories: \qty{<5}{\minute}, \qtyrange{5}{30}{\minute}, \qtyrange{30}{120}{\minute}, and \qty{>2}{\hour}.
These groupings were based on experimental iterations to maximise the association between LVI features and seizure burden.

For these sets of clinical evaluation experiments, we applied the Kruskal--Wallis $H$-test to first determine if there was any significant separation among the groups.
We used Epsilon-squared ($\varepsilon^2 = H/(n-1)$) as an effect size measure, which evaluated the proportion of variance of the ranked LVI feature explained by the groups.
Because the grades in all these experiments are ordinal, we also applied Spearman's rank correlation coefficient ($\rho$) to quantify the monotonic dependency of grades on LVI features.
Statistical analysis was performed in \textit{Python} (v3.13) using \textit{SciPy} (v1.14).
Statistical significance was set at $p<0.05$.
Confidence intervals were generated through a bootstrapped process with \num{10000} iterations and sampling across neonates.

\section{Results}
\label{sec:results}

\subsection{Part A: LVI detection performance}
\label{sec:lvi-detect-perf}

Across the development and validation sets, \qty{654}{\hour} of per-channel EEG data were reviewed and annotated for LVIs (Table~\ref{tab:train_test_set}).
In total, \num{51162} LVIs were annotated: \num{36580} in the development set and \num{14582} in the held-out validation set.
The ratio of LVIs to non-LVI EEG was approximately 1:4 in the development set and 1:2 in the validation set (Table~\ref{tab:lvi_annotation_intervals}).
Inter-rater agreement for LVI annotation between the two human readers on the validation set was high, with a Cohen's $\kappa$ of 0.82 (95\% CI: 0.74 to 0.87).

The SVM and ConvNeXt ML methods consistently outperformed the quantitative features in the 10-fold
cross-validation on the development set (see Fig.~\ref{fig:ibi_xv_results} in Appendix).
Between the two ML methods, ConvNeXt marginally outranked SVM on almost all metrics.
For example, on the concatenated results, both methods achieved an area under the receiver operating characteristic curve (AUC) of 0.977, while ConvNeXt produced a higher Matthews correlation coefficient (MCC) compared to the SVM (0.802 versus 0.794).

\begin{figure}[!ht]
  \centering
  \includegraphics[width=0.99\textwidth]{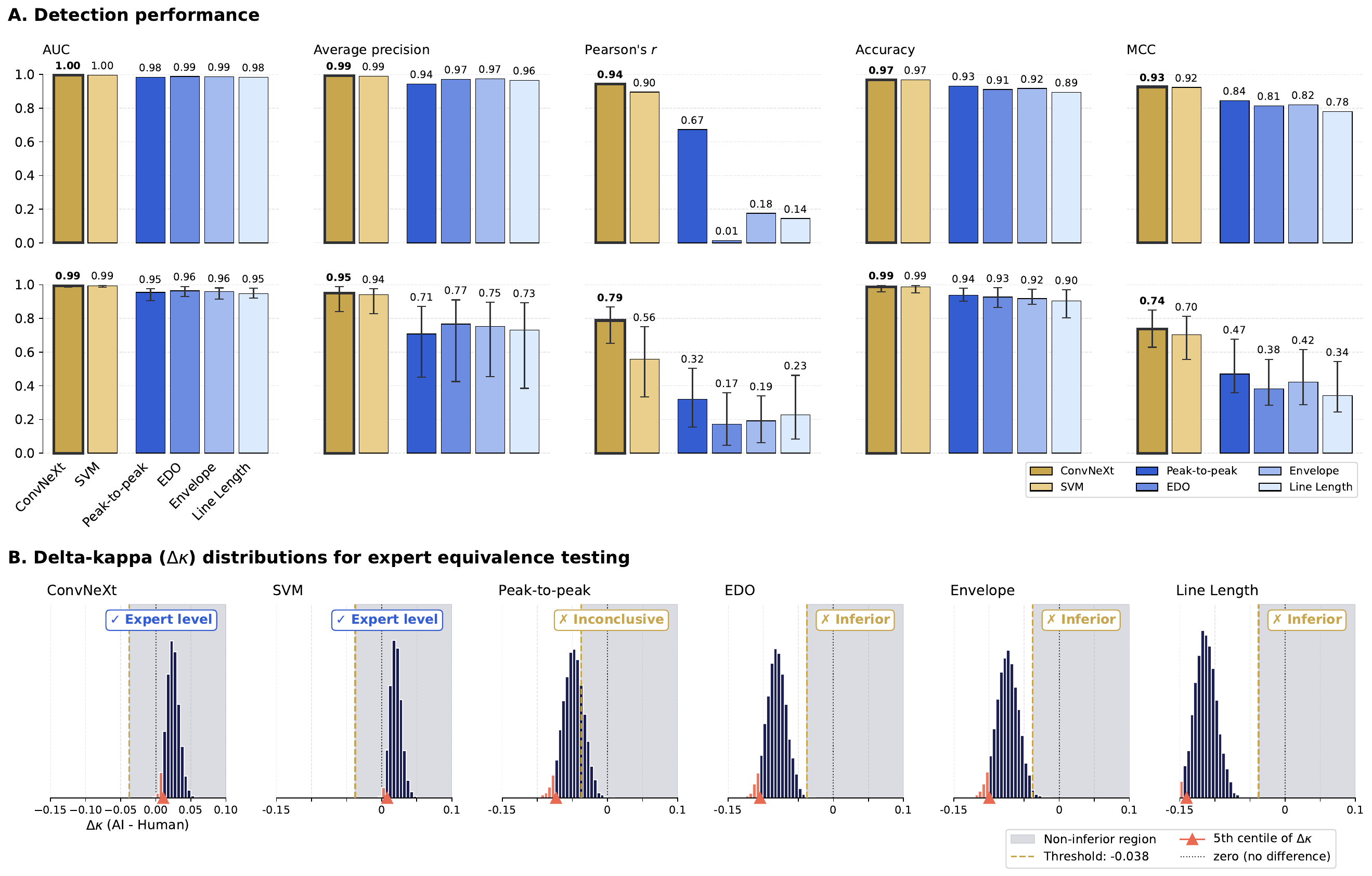}
  \caption{Performance of the low-voltage interval (LVI) detectors on the held-out validation set.
A: Performance metrics calculated from annotations concatenated across all EEG records (top) and on a per-neonate basis (bottom), with medians and interquartile ranges shown for the latter.
Consensus annotations from two human experts were used.
B: Expert-level equivalence testing.
The $\Delta \kappa$ distribution is the difference in agreement (Cohen's $\kappa$) between the detector and each human annotation compared with the agreement between the two human experts.
Positive values of $\Delta \kappa$ indicate an increase in agreement between the detection method (labelled as AI here) and experts compared with human-level agreement.
Both ConvNeXt and SVM detectors reach expert-level agreement.
Key: SVM, support vector machine; EDO, energy derivative operator.
}\label{fig:cnx_expert_testing}
\end{figure}

The trend was similar on the held-out validation set: the SVM and ConvNeXt methods outperformed the quantitative features, with ConvNeXt outranking the SVM.
Performance metrics are illustrated in Fig.~\ref{fig:cnx_expert_testing}A with full details in Tables~\ref{tab:heldout_concat_results} and \ref{tab:heldout_perbaby_results}.
For example, on the unanimous annotation for the concatenated performance, both methods---again---achieved the same AUC (0.996) with a marginally higher MCC for ConvNeXt compared to the SVM (0.926 versus 0.923).

The expert-level equivalence testing reinforced this finding, with both SVM and ConvNeXt methods comfortably achieving expert-level equivalence ($p<0.001$) whereas the quantitative features did not ($p>0.05$).
As Fig.~\ref{fig:cnx_expert_testing}B shows, $\Delta \kappa$ estimates for both methods were positive: 0.020 (95\% CI: 0.008--0.034) for the SVM and 0.024 (95\% CI: 0.011--0.040) for the ConvNeXt.
Again, the ConvNeXt had a slight edge over the SVM with a larger (positive) $\Delta \kappa$.
See Table~\ref{sec:expert_testing} for $\Delta \kappa$ values for all methods.
For this reason, we use the ConvNeXt in all application experiments in the following Part B section.

\subsection{Part B: Clinical Relevance and Biomarker Potential}
\label{sec:clin-eval-exper}

We summarised LVI activity for each epoch using the six quantitative features described in Section
\ref{sec:app_lvi}.
We tested the accuracy of these LVI features using the validation set by comparing LVI features generated from the model's predictions with LVI features generated from the experts' annotations.
For each feature, we calculated the baseline inter-rater variability as the absolute difference between the two human experts and the ML model error as the absolute difference between the model's estimate and the experts' average.
Then we resampled the set using a bootstrap clustered over neonates and calculated a one-sided empirical $p$-value as the proportion of samples where ML model error was greater than or equal to the human inter-rater error.

Four of the six LVI features demonstrated non-inferiority or superiority to expert inter-rater variability ($p<0.05$).
For two features, namely the maximum and median duration, we failed to reject the null hypothesis.
We did find, however, that the magnitudes of these errors were small and therefore unlikely to be clinically meaningful: \qty{0.100}{\second} (95\% CI: -0.686 to 0.856 s) for maximum duration and -\qty{0.172}{\second} (95\% CI: -0.188 to 0.594 s) for median duration.
Full details are provided in Table~\ref{tab:lvi_feature_differences}.

For the first LVI application, the six features were computed for the 169 \qty{1}{\hour} epochs ($n=53$
neonates) from the open-access EEG dataset with the four grades of background abnormalities \cite{OToole2023}.
All features showed good separation across the grades; see Fig.~\ref{fig:eeg_hie_grades}A and Table~\ref{tab:lvi_app_eeg_grades} in the Appendix.
Effect size ($\varepsilon^2$) and Spearman's correlation coefficient ($\rho$) values for all six features were significant ($p<0.001$).
The median-duration LVI feature achieved the strongest separation across grades, with $\rho=0.821$
and $\varepsilon^2 = 0.676$.

\begin{figure}[!ht]
  \centering
  \includegraphics[width=0.99\textwidth]{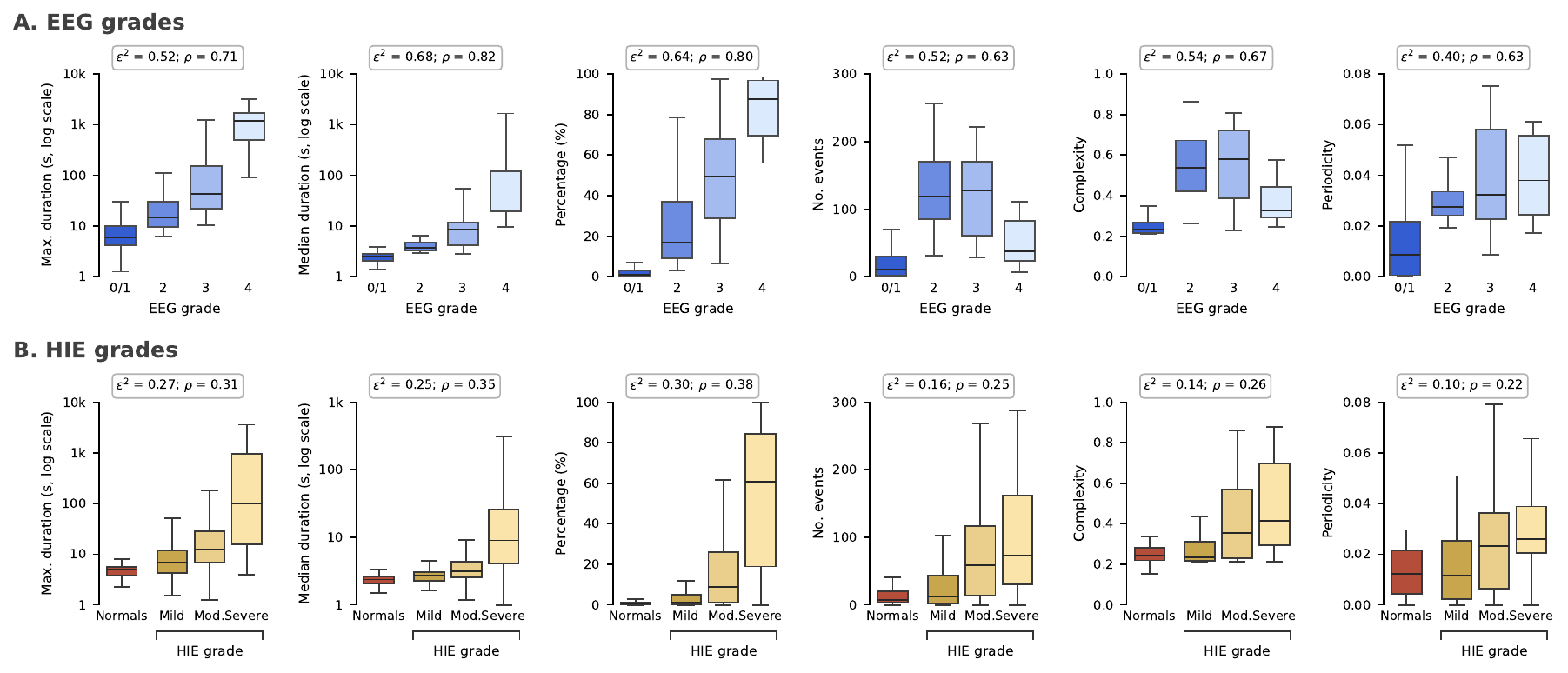}
  \includegraphics[width=0.99\textwidth]{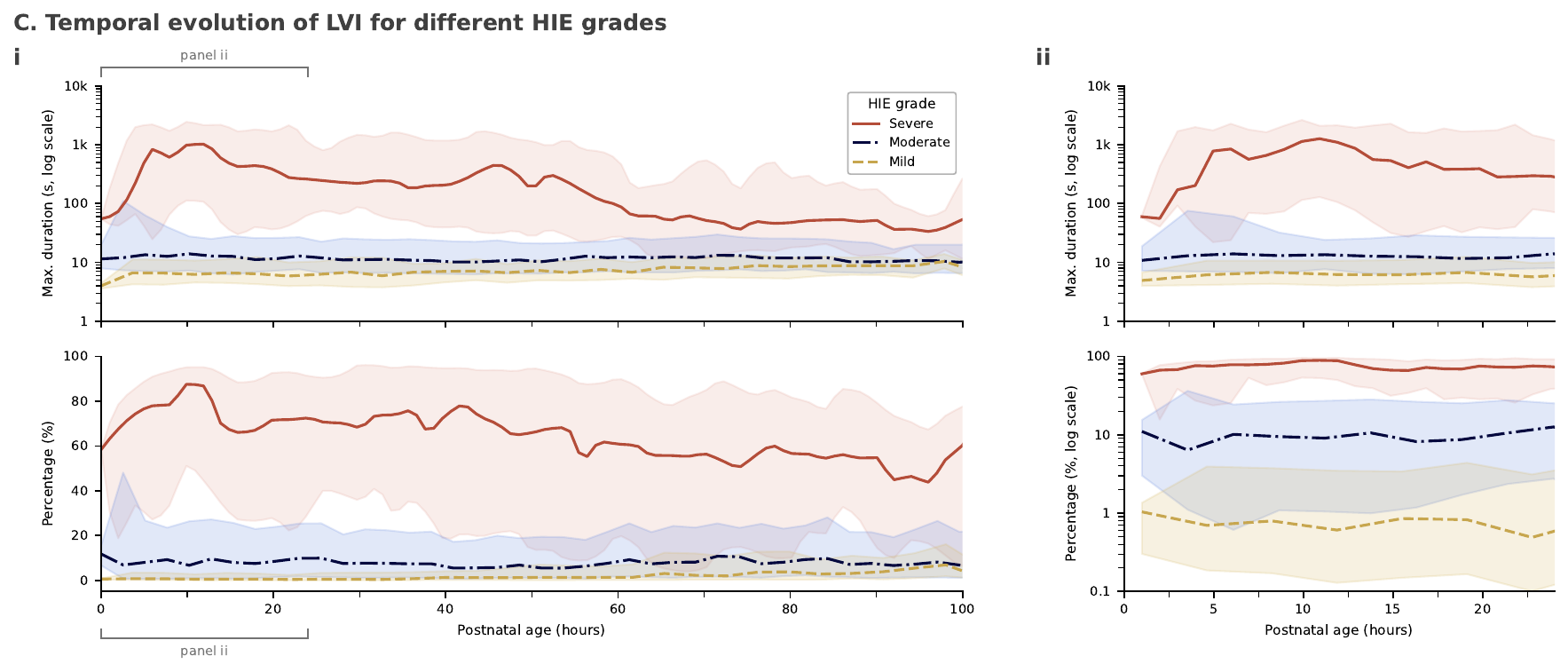}
  \caption{Examples of clinical relevance of LVIs.
Panels A and B: 6 LVI features generated from \qty{1}{\hour} of EEG with effect-size measure epsilon-squared ($\varepsilon^2$) and Spearman's rank correlation coefficient ($\rho$).
Panel A: LVI features are generated for 4 different EEG grades of background: normal/mild (0/1), moderate (2), and major (3) abnormalities, and inactive EEG (4).
Panel B: HIE grades are the based on the modified Sarnat score, in addition to EEG from healthy neonates (normals) for comparison.
Panel C: Two LVI features generated across time for the long-duration EEG recordings by generating features on a 1 h epoch shifted by \qty{15}{\minute}.
Panel C(ii): the first 24 hours of the plot in panel C(i) in more detail.
}
  \label{fig:eeg_hie_grades}
\end{figure}

For the second LVI application, we examined separation among the clinical HIE grades.
A total of 223 neonates had a confirmed diagnosis of HIE from the clinical evaluation dataset.
Four were excluded as the grade---either mild, moderate, or severe---was missing.
The random-sampling process, described in Section \ref{sec:app_lvi}, produced 571 \qty{1}{\hour} epochs from the remaining 219 neonates.
A further 48 \qty{1}{\hour} epochs from 48 healthy neonates were added to this analysis to provide a reference baseline.
After computing the LVI features, we again found a significant separation ($p<0.001$), albeit to a lesser extent than that for the EEG grades.
See Fig.~\ref{fig:eeg_hie_grades}B and Table~\ref{tab:lvi_app_eeg_grades} in the Appendix for details.
The LVI percentage feature produced the strongest separation across the four grades, with $\rho=0.384$ and $\varepsilon^2 = 0.297$.

We also tracked LVI features across the first 100 hours after birth.
Two LVI features, namely the maximum duration and LVI percentage, were computed on \qty{1}{\hour} epochs with a 25\% overlap.
Over \qty{13000} hours of multichannel EEG were analysed with more than 30 million LVI events detected by the algorithm.
Fig.~\ref{fig:eeg_hie_grades}C illustrates this temporal trajectory, revealing a slow but consistent decrease in maximum duration over time after approximately 10 hours for the severe HIE group.
In contrast, features for the mild and moderate grades appeared relatively stable across time.
As the first 24 hours are an important period in the neurocritical management of these at-risk
neonates, this trajectory is also presented in a separate plot (Fig.~\ref{fig:eeg_hie_grades}C-ii),
where a log scale of LVI percentage is used to highlight the separation between the three HIE
grades.

Our third application experiment was to analyse neonates with seizures.
For this, we used the subset of our clinical evaluation dataset (Table~\ref{tab:evaluation_set}) that were reviewed and annotated for seizures ($n=291$).
For the seizure prediction task, we extracted the first \qty{1}{\hour} epoch from each EEG recording and computed the LVI features.
We excluded 88 neonates with seizure onset \qty{<4}{\hour} after start of recording, resulting in a median seizure onset relative to the start of the EEG of \qty{8.4}{\hour} (IQR: \qtyrange{6.1}{11.4}{\hour}) with a range of \qtyrange{4.0}{38.3}{\hour} ($n=203$).
We then examined whether the LVI features in the first hour of EEG recording were predictive of which neonates would subsequently have seizures.
All features showed significant separation between classes ($p<0.001$), with LVI percentage as the leading discriminating feature with an AUC of 0.872 and $\rho=0.475$.
Fig.~\ref{fig:sz_lvi_features}A and Table~\ref{tab:lvi_app_sz} in the Appendix present performance for all features for the seizure prediction task.

\begin{figure}[!ht]
  \centering
  \includegraphics[width=0.99\textwidth]{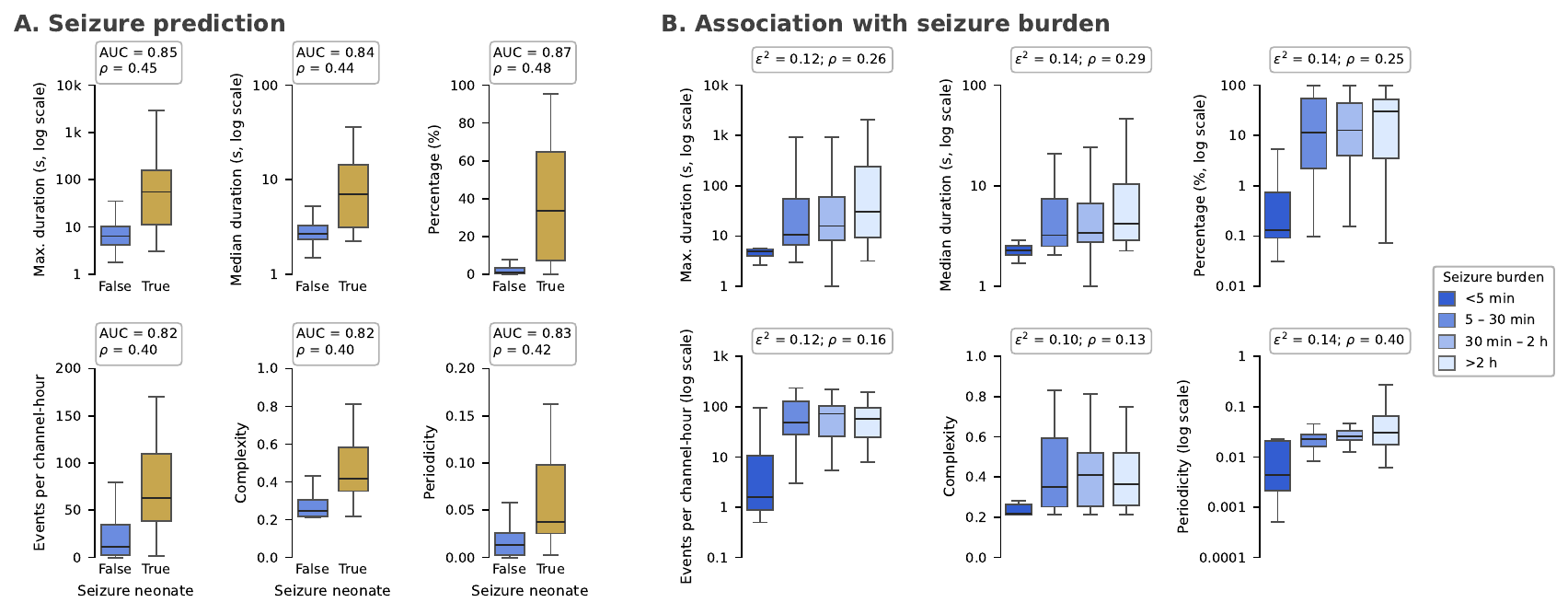}
  \caption{Third example of clinical relevance of LVIs.
A: LVI features generated from the first hour of EEG used to separate neonates with seizures from neonates without seizures.
Area under the receiver operating characteristic curve (AUC) and Spearman's rank correlation coefficient ($\rho$) used as discrimination measures.
B: LVI features generated over the whole EEG recording for those neonates with seizures.
Association between LVI features and seizure burden measured using $\rho$ and discrimination of 4 categories of seizure burden measured with epsilon-squared ($\varepsilon^2$).
} \label{fig:sz_lvi_features}
\end{figure}

In the fourth and final application experiment, we evaluated the association of LVI activity with seizure burden.
For those neonates with seizure in our dataset (\newtext{$n=121$}), we calculated seizure burden by summing the total time of all seizures for each neonate.
We computed the features for \qty{1}{\hour} epochs with a 75\% overlap and compared them with seizure burden using correlation ($\rho$) to quantify the relation between the two continuous variables.
We found that five of the six features had a significant ($p<0.05$) but weak association with seizure burden, with the highest correlation observed for the LVI periodicity feature ($\rho=0.398$).
We then stratified seizure burden into four groups based on duration of burden, as detailed in Section~\ref{sec:app_lvi}, and again found a weak but significant ($p<0.05$) effect size for five of the six LVI features across the four categories of seizure burden.
For the stratified version, the LVI periodicity feature had the largest effect size ($\varepsilon^2=0.138$); see Fig.~\ref{fig:sz_lvi_features}B and Table~\ref{tab:lvi_app_sz} for more details.

\subsubsection{Robustness of LVI biomarkers}

\begin{figure}
  \centering
  \includegraphics[width=1.00\textwidth]{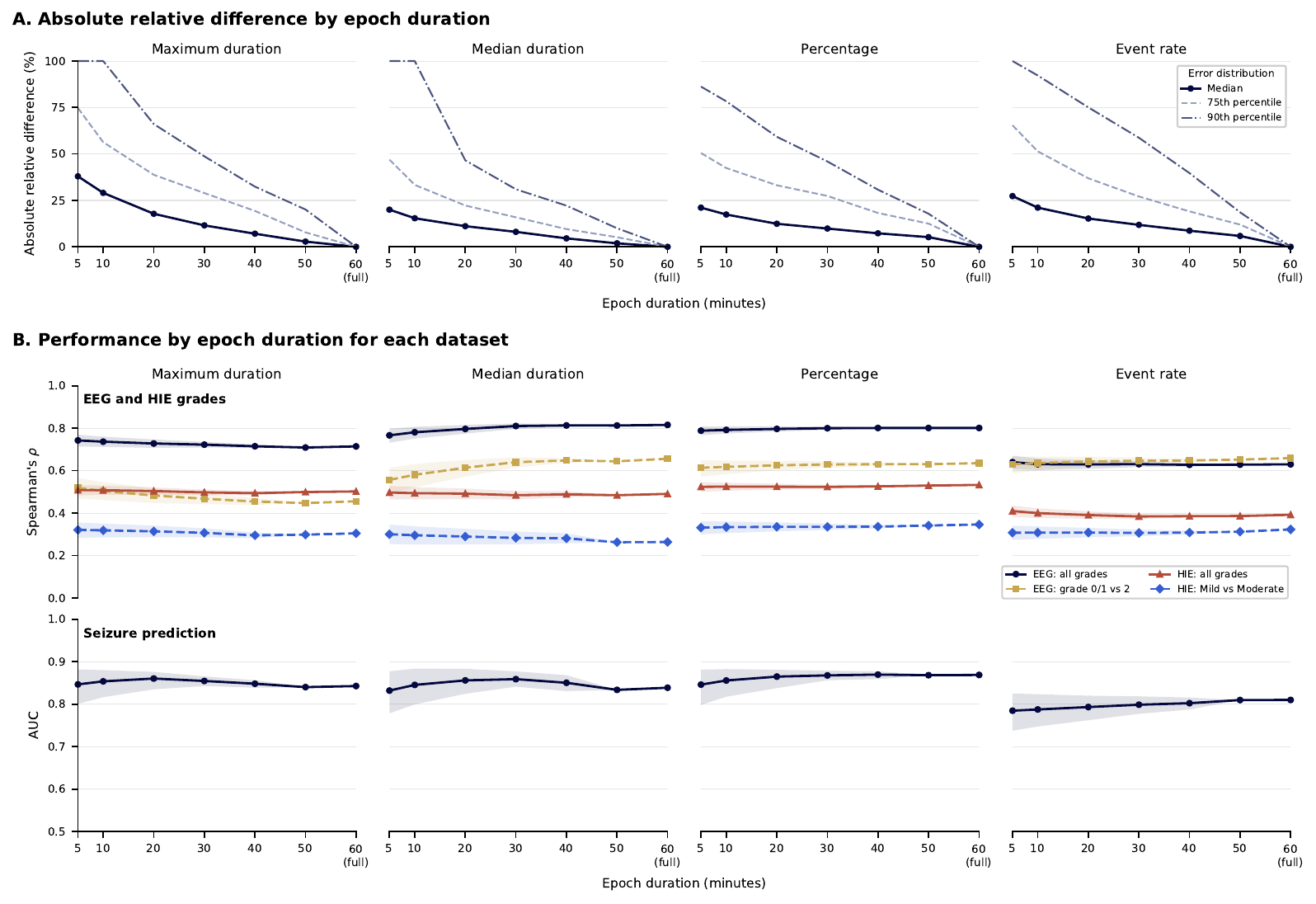}
  \caption{Effect of epoch duration on LVI features and biomarker performance.
A: Difference in feature values for epochs $<$1 hour relative to values calculated on the full 1-hour epoch.
Short-duration epochs are generated within the full-duration epoch at a 75\% overlap rate.
Absolute relative error is computed by comparing each short-duration to the full-duration epoch.
The resulting error distribution is outlined with percentiles at the 50th (median), 75th, and 90th mark.
This analysis uses the open-access EEG grades dataset.
B: Performance metrics, either Spearman's rank correlation coefficient or area under the receiver operating characteristic curve (AUC), for shorter-duration epochs compared to the full-duration epoch.
This analysis is applied to three applications: separating EEG grades, separating HIE clinical
grades, and predicting seizures.
Performance metrics are calculated using one short-duration epoch, randomly sampled from each
full-duration epoch and repeated \num{10000} times.
Median values are represented as markers and the 95\% resampling interval as the shaded region.
}
  \label{fig:short_epochs_analysis}
\end{figure}

To test the sensitivity of the LVI features to epoch duration, we first compare feature values calculated on shorter-duration epochs and then perform the preceding LVI application analysis with these shorter epochs.
As expected, we find a relatively large difference in feature values for the shorter epochs compared to the full 1-hour epoch, as illustrated in Fig.~\ref{fig:short_epochs_analysis}A.
The shorter the epoch, the larger the absolute relative difference from the full-duration epoch.
Nonetheless, predictive performance of the LVI features remains relatively consistent regardless of epoch duration, as illustrated in Fig.~\ref{fig:short_epochs_analysis}B where we find no discernible change.
This robustness to epoch duration is most prominent in the EEG and HIE grading tasks; the seizure prediction task shows greater variability for shorter-duration epochs (Fig.~\ref{fig:short_epochs_analysis}B), although the median AUC estimates at 5 or \qty{10}{\minute} epochs are strikingly close to the AUC obtained using the 1-hour epoch.

A large proportion of LVIs are short in duration, for example \qty{<6}{\second}, as indicated in Fig.~\ref{fig:ibi_duration_analysis}A.
To test the clinical relevance of these short events in quantitative analysis, we performed a two-part analysis using the preceding biomarker tasks.
First, we removed short-duration LVIs by systematically increasing a minimum duration threshold and evaluating the impact on feature performance across the range of tasks.
Fig.~\ref{fig:ibi_duration_analysis}B shows no obvious trend of improved performance when shorter-duration LVIs are removed.
Second, we considered only short-duration LVIs by systematically increasing a maximum duration threshold, again evaluating the impact on performance.
We found that restricting analysis to short-duration LVIs still produces non-zero performance, even for LVIs $\leq$\qty{4}{\second} (see Fig.~\ref{fig:ibi_duration_analysis}C).
Together, these results suggest that short-duration LVIs may themselves be informative, or at the very least, their removal does not improve analysis.

Enforcing a lower limit on IBI duration, for example ignoring IBIs \qty{<6}{\second}, is a requirement for human EEG review but is not necessary for a computer-based approach that can analyse all events within an epoch.
Not requiring a predefined absolute minimum threshold ensures the robustness of the LVI features, particularly given their sensitivity to epoch duration.
We also find that the amplitude of the EEG within LVIs decreases as LVI duration increases, as illustrated in Appendix Fig.~\ref{fig:ibi_duration_amplitude}.
This is consistent with our concept of the LVI as a super-set, including IBIs associated with tracé discontinu (\qty{<25}{\micro\volt}), burst suppression (\qty{<5}{\micro\volt}), and even isoelectric activity (\qty{<2}{\micro\volt}).

\section{Discussion}
\label{sec:discussion}

Both ML models achieved expert-level equivalence in detecting LVIs.
This success was driven by a large dataset of 654 hours of per-channel EEG with 51k individual LVI annotations.
This represents an increase of two orders of magnitude over the previous largest annotated dataset of 6 hours of per-channel IBI annotation in preterm EEG \cite{O’Toole2017}.
This unprecedented volume of data allowed us to build robust models that generalise well.
In addition to model performance, automated detection of LVIs demonstrates potential as a biomarker for brain injury across an EEG dataset from 444 neonates.
Objective and repeatable quantitative LVI features are able to stratify EEG background grades as well as clinical grades of HIE, and accurately predict which neonates will later have seizures.
These findings together demonstrate the clinical relevance and importance of LVIs in characterising pathological EEG activity related to neonatal brain injury.

Exceptions will inevitably arise when applying strict EEG definitions, in our case that an LVI is
any activity that remains \qty{<25}{\micro\volt} for $\geq$\qty{2}{\second}.
The experienced EEG reader must exercise discernment and nuance to correctly interpret the complex
EEG signal before annotating an event.
This is illustrated by the performance of the peak-to-peak quantitative feature, which strictly implements the rule yet fails to match human accuracy.
The ML approach, in contrast to rigid quantitative features \cite{Palmu2010a,Koolen2014,Matic2016,Mader2024}, learns the essential characteristics from the expert annotations, enabling it to capture the implicit rules and contextual assumptions embedded in human review.

Compared to the SVM, the ConvNeXt model had marginally better performance on almost all metrics, although it is likely that the difference is not clinically meaningful.
This differs from many other ML applications where deep-learning models typically outperform feature-based ML models.
In our case, the similarity between the methods is likely because of the simplicity of the task: LVIs may be more uniform and less complex than, for example, seizure waveforms.
Regardless of performance, there are some distinct advantages to the ConvNeXt model.
We found that the SVM output, an unbounded output compressed by a sigmoid function to represent a probability value, was noisy compared to the ConvNeXt probability, as illustrated in Fig.~\ref{fig:overview_lvis}B-ii.
Also, the inference pipeline for the ConvNeXt is simpler, with many popular libraries supporting highly-optimised implementation of convolutional models.
Conversely, the qEEG features required for the SVM model are not standard and lack widespread library support.
Neither method requires a large computational load, meaning both could be run in real-time using a standard PC or laptop.
Because the ConvNeXt model is small compared to other deep-learning models, with \num{<40}k parameters, it does not require specialised hardware acceleration and therefore could be integrated into cot-side monitoring.

We used six features to characterise the binary LVI detection mask.
Maximum and median duration, percentage, and number of events are simple, intuitive measures that have been used elsewhere \cite{OToole2017,Koolen2014}.
To account for the temporal dynamics of LVI events, we included complexity and periodicity.
Although these two features were not among the highest-ranked features in the clinical application experiments, they quantify information that the other features do not capture, and therefore may be useful, for example, when combined in an ML algorithm.
While further features could represent more information encoded in the mask, a small and
interpretable feature set has its own advantages.

The clinical neurophysiology approach to summarising IBI patterning is to estimate the maximum duration of IBI events.
This approach has inherent limitations, however.
The EEG recording must be of sufficient duration to produce a reliable estimate, and accurate visual assessment of a maximum event can be challenging, particularly with longer EEG records.
Quantitative features are, in contrast, objective and repeatable.
Although we find that these features are sensitive to epoch duration (Fig.~\ref{fig:short_epochs_analysis}A), when the same duration epoch is used in discriminatory analysis, performance is on-par with that from the longer duration epochs (Fig.~\ref{fig:short_epochs_analysis}B).
This indicates that shorter-duration epochs, for example a \qty{10}{\minute} epoch, may be sufficient to summarise the clinically important LVI activity within the full 1-hour epoch.
Absolute thresholds for LVI features are best determined by quantitative data-driven approaches, such as machine-learning, rather than defining values based on clinical IBI thresholds or limited analysis; regardless, epoch duration of analysis should be considered and consistent under analysis.

LVI features show remarkably good discriminatory power across the EEG grades.
For example, Spearman's correlation across the four grades for median LVI duration was $\rho=0.82$ ($p<0.001$).
Because classification of background EEG activity is an important marker of real-time brain health, there has been a long history of automated classification methods for this task \cite{Stevenson2012,Matic2014,Matic2016,Raurale2021,Moghadam2022,OtextquotesingleSullivan2023,Twomey2025}.
One grading system in particular, known as the Brain State of the Newborn (BSN), has shown the capacity to stratify HIE clinical grades, early clinical markers, and longer-term neurodevelopmental outcomes \cite{Montazeri2024,Tuiskula2024,Kota2024,Lagace2024,GonzalezTamez2025,Cornet2025}.
This extensive validation on different independent datasets has been possible through the provision of an open-access platform to upload EEG and return the processed results \cite{Dilena2024}.
Yet it can be difficult to know what the end-to-end ML methods are learning in training \cite{Raurale2021,Moghadam2022,OtextquotesingleSullivan2023,Twomey2025}.
These methods learn from long segments of EEG, ranging from 1 to 60 minutes, which may contain significant dynamic EEG activity.
They are also trained with relatively small datasets given the complexity of the task, $n<55$ for \cite{Raurale2021,Moghadam2022,Twomey2025} with a larger set of $n=181$ used in \cite{OtextquotesingleSullivan2023}.
Therefore it is possible that these algorithms are learning from the broader and more obvious EEG characteristics such as amplitude rather than the more subtle EEG patterns which include transient waveforms.
Our findings support this theory that LVI activity could dominate the end-to-end learning process for these methods \cite{Raurale2021,Moghadam2022,OtextquotesingleSullivan2023,Twomey2025}.
Given that amplitude and continuity are common factors in many grading systems \cite{Walsh2011}, it is correct and essential that ML systems capture these characteristics.
It remains to be seen, however, whether ML methods are able to account for all types of abnormalities in the EEG, including the more subtle waveforms rather than just summarising LVI activity.

We found significant separation for all LVI features across the three clinical grades of HIE and the normal cohort, although the discriminatory power is less than that for the EEG grades.
LVI percentage, the highest ranked feature, had a correlation of $\rho=0.38$ ($p<0.001$).
The grade of HIE is derived from the modified Sarnat score, which is based on the antenatal history
of the hypoxic-ischaemic event and early neurological examination, and then later confirmed with
changes in EEG background or MRI that are consistent with HIE \cite{Pavel2022}.
Sarnat scores can change over time with evolving injury and therefore may not be representative of
the analysed EEG epoch.
These \qty{1}{\hour} epochs were selected at random time points across the recording duration and may reflect evolving injury.
LVI features for the mild and moderate HIE grades did, however, appear relatively stable over the first \qty{100}{\hour} of EEG (Fig.~\ref{fig:eeg_hie_grades}C-i), and in particular the first 24 hours (Fig.~\ref{fig:eeg_hie_grades}C-ii), where this information could be the most clinically useful.
The percentage of LVI, for example, shows stable separation across the 24 hours for all three grades, indicating the capability for LVI measures to track development or recovery of EEG activity in real time.

LVI features were able to discriminate between neonates that would subsequently have seizures and
those who did not, with an AUC of 0.88 for median duration and LVI percentage.
A notable finding given that the median (IQR) time to the first seizure from the analysed epoch was 8 (6 to 11) hours.
Strikingly, this level of performance is comparable with ML methods that are trained specifically for this task \cite{Pavel2022,Bernardo2024}.
This provides more evidence on the clinical relevance of the LVI events through automated detection and quantification.

The association of LVI activity across the entire EEG recording with seizure burden was significant but weak.
For example, maximum duration had a significant correlation with seizure burden ($\rho=0.27$) and there was a difference among the four categories of burden ($\varepsilon^2=0.15$).
Given the many potential confounding factors, detecting even a noisy and weak association remains a positive finding.
LVI features were calculated across the whole EEG recording; in some instances, the seizures may have stopped and the EEG recovered.
Anti-seizure medication may have stopped or greatly reduced seizure burden but the EEG could still be discontinuous.
Other clinical factors, such as therapeutic hypothermia, may produce a dissociation between LVI and seizure burden.

While this study has several strengths, it is not without limitations.
First, the validation EEG dataset is from the same centre as the development (training) dataset.
Ideally, the validation dataset would come from a different centre to assure generalisability of the models.
However, there is no overlap in EEG readers between the development and validation sets, and the validation set is a mix of heterogeneous studies.
Also, the validation set is relatively large with \num{14582} annotated events.
Second, our validation dataset was selected to be free from major artefacts.
Therefore performance may degrade in the presence of significant artefacts and a pre-processing artefact detector may be required \cite{OtextquotesingleSullivan2023}.
However, we did find good discrimination on the HIE and normal datasets and on the seizure prediction task where the epochs were not screened for artefact.
Such screening may even further improve these evaluations.
Third, we do not present recommended thresholds for the LVI features to indicate a clinical concern.
Rather than enforcing rigid discrete categories, we present these features as a continuum and suggest that wider clinical validation is required before setting dichotomous cut-offs, as different applications may require tailored thresholds.
Fourth, our definition of an LVI differs from that of an IBI.
IBIs have different definitions depending on the context.
In tracé alternant, IBIs are between 25 and \qty{50}{\micro\volt}, \qty{<5}{\micro\volt} in burst suppression, and \qty{<25}{\micro\volt} in tracé discontinu and nearly-continuous activity; whereas our LVI definition is activity \qty{<25}{\micro\volt} regardless of vigilance state or background activity.
It is therefore not certain if the clinical evaluation results would generalise to traditional IBI events.

In summary, we present an automated method to detect LVIs with performance indistinguishable from that of an expert.
Although the detection of LVIs may seem like a trivial exercise, given the strict and simple definition, we find that an ML approach with a large set of annotations is required to attain expert-level detection and that simple feature-based methods fail to achieve comparable results.
We thoroughly validate automated LVI detection on a large dataset ($n=444$) and show
its clinical potential as a biomarker of neonatal encephalopathy.
LVI detection may prove to be an important component for clinical decision support or an input module for an application-specific detection or prediction algorithm.
Automated analysis has many advantages in assisting with the challenges of neonatal EEG review, as it provides an objective, repeatable, and quantifiable measure.

\section{Data Availability}

The benchmark EEG dataset with background grades is freely available at \PaperDatasetLink.
We do not have permission to share the rest of the EEG data used in this study because of the terms of the data licensing agreement and the terms of the informed consent provided by parents or guardians, as approved by the ethics committee.
Code to generate the quantitative features and machine-learning models is available at
\url{https://github.com/cergenx/lvi-detection} (commit 48b2f19).

\section{Acknowledgements}
This research was funded by an Enterprise Ireland Disruptive Technology Innovation Fund Award (DT 2023 435).
Enterprise Ireland played no role in study design, data collection, analysis and interpretation of
data, or the writing of this manuscript.
We gratefully acknowledge Kylie Hale for her contribution to the LVI annotation work used in this study.

\section{Competing Interests}
All authors are affiliated with CergenX Ltd, a University College Cork spin-out company developing
neuromonitoring technologies for newborns: G.B.B. and S.G. are co-founders; R.H., A.L., and
J.O.T. are employees; and S.M. and S.V. were independent consultants.

\FloatBarrier
\PaperReferencesHeading
\bibliographystyle{ieeetr}
\bibliography{references}

@article{Schendel2014,
  author =        {Schendel, Diana},
  journal       = {Obstetrics \& Gynecology},
  number =        {4},
  pages =         {896--901},
  publisher =     {Lippincott},
  title =         {Executive summary: neonatal encephalopathy and
                   neurologic outcome, Report of the {American College
                   of Obstetricians and Gynecologists' } task force on
                   neonatal encephalopathy},
  volume =        {123},
  year =          {2014},
  doi =           {https://doi.org/10.1097/01.AOG.0000445580.65983.d2},
}

@article{Lawn2005,
  author =        {Lawn, Joy E and Cousens, Simon and Zupan, Jelka},
  journal =       {The Lancet},
  month =         mar,
  number =        {9462},
  pages =         {891--900},
  publisher =     {Elsevier BV},
  title =         {4 million neonatal deaths: When? Where? Why?},
  volume =        {365},
  year =          {2005},
  doi =           {10.1016/s0140-6736(05)71048-5},
  issn =          {0140-6736},
}

@article{Tan2025,
  author =        {Tan, Shujun and Mageziyi, Shala and Long, Li and
                   Dilimulati, Naiyiman and Hui, Zhang and
                   Rejiafu, Nuerya},
  journal =       {Frontiers in Public Health},
  month =         oct,
  publisher =     {Frontiers Media SA},
  title =         {Analysis of global, regional, and national burdens of
                   neonatal encephalopathy from 1990 to 2021: insights
                   from the Global Burden of Disease Study 2021},
  volume =        {13},
  year =          {2025},
  doi =           {10.3389/fpubh.2025.1627448},
  issn =          {2296-2565},
}

@article{Moghadam2022,
  author =        {Moghadam, Saeed Montazeri and Airaksinen, Manu and
                   Nevalainen, Päivi and Marchi, Viviana and
                   Hellström-Westas, Lena and Stevenson, Nathan J and
                   Vanhatalo, Sampsa},
  journal =       {The Lancet Digital Health},
  month =         dec,
  number =        {12},
  pages =         {e884--e892},
  publisher =     {Elsevier BV},
  title =         {An automated bedside measure for monitoring neonatal
                   cortical activity: a supervised deep learning-based
                   electroencephalogram classifier with external cohort
                   validation},
  volume =        {4},
  year =          {2022},
  doi =           {10.1016/s2589-7500(22)00196-0},
  issn =          {2589-7500},
}

@article{OtextquotesingleSullivan2023,
  author =        {Mark E. O'Sullivan and Gordon Lightbody and
                   Sean R. Mathieson and William P. Marnane and
                   Geraldine B. Boylan and John M. O'Toole},
  journal =       {Expert Systems with Applications},
  month =         mar,
  pages =         {118917},
  publisher =     {Elsevier {BV}},
  title =         {Development of an {EEG} artefact detection algorithm
                   and its application in grading neonatal
                   hypoxic-ischemic encephalopathy},
  volume =        {213},
  year =          {2023},
  doi =           {10.1016/j.eswa.2022.118917},
}

@article{Twomey2025,
  author =        {Twomey, Leah and Gomez, Sergi and Popovici, Emanuel and
                   Temko, Andriy},
  journal =       {Sensors},
  month =         may,
  number =        {10},
  pages =         {3007},
  publisher =     {MDPI AG},
  title =         {Long-Term Neonatal {EEG} Modeling with {DSP} and {ML}
                   for Grading Hypoxic–Ischemic Encephalopathy Injury},
  volume =        {25},
  year =          {2025},
  doi =           {10.3390/s25103007},
  issn =          {1424-8220},
}

@article{Montazeri2024,
  author =        {Montazeri, Saeed and Nevalainen, Päivi and
                   Metsäranta, Marjo and Stevenson, Nathan J. and
                   Vanhatalo, Sampsa},
  journal =       {Clinical Neurophysiology},
  pages =         {68--76},
  publisher =     {Elsevier BV},
  title =         {Clinical outcome prediction with an automated {EEG}
                   trend, {Brain State of the Newborn}, after perinatal
                   asphyxia},
  volume =        {162},
  year =          {2024},
  doi =           {10.1016/j.clinph.2024.03.007},
  issn =          {1388-2457},
}

@article{Tuiskula2024,
  author =        {Tuiskula, Anna and Pospelov, Alexey S. and
                   Nevalainen, Päivi and Montazeri, Saeed and
                   Metsäranta, Marjo and Haataja, Leena and
                   Stevenson, Nathan and Tokariev, Anton and
                   Vanhatalo, Sampsa},
  journal =       {Pediatric Research},
  month         = jan,
  publisher =     {Springer Science and Business Media LLC},
  title =         {Quantitative {EEG} features during the first day
                   correlate to clinical outcome in perinatal asphyxia},
  year          = {2025},
  doi =           {10.1038/s41390-024-03235-y},
  issn =          {1530-0447},
  volume        = {97},
  number        = {1},
  pages         = {261--267},
}

@article{Kota2024,
  author =        {Kota, Srinivas and Kang, Shu and Liu, Yu-Lun and
                   Liu, Hanli and Montazeri, Saeed and Vanhatalo, Sampsa and
                   Chalak, Lina F.},
  journal =       {Pediatric Research},
  number =        {3},
  pages =         {685--694},
  publisher =     {Springer Science and Business Media LLC},
  title =         {Prognostic value of quantitative {EEG} in early hours
                   of life for neonatal encephalopathy and
                   neurodevelopmental outcomes},
  volume =        {96},
  year =          {2024},
  doi =           {10.1038/s41390-024-03255-8},
  issn =          {1530-0447},
}

@article{Lagace2024,
  author =        {Lagacé, Micheline and Montazeri, Saeed and
                   Kamino, Daphne and Mamak, Eva and Ly, Linh G. and
                   Hahn, Cecil D. and Chau, Vann and Vanhatalo, Sampsa and
                   Tam, Emily W. Y.},
  journal =       {Annals of Clinical and Translational Neurology},
  month =         nov,
  number =        {12},
  pages =         {3267--3279},
  publisher =     {Wiley},
  title =         {Automated assessment of {EEG} background for
                   neurodevelopmental prediction in neonatal
                   encephalopathy},
  volume =        {11},
  year =          {2024},
  doi =           {10.1002/acn3.52233},
  issn =          {2328-9503},
}

@article{GonzalezTamez2025,
  author =        {Gonzalez-Tamez, Karla and Montazeri, Saeed and
                   Ågren, Johan and Vanhatalo, Sampsa and
                   Hellström-Westas, Lena},
  journal =       {Pediatric Research},
  number =        {2},
  pages =         {710--717},
  publisher =     {Springer Science and Business Media LLC},
  title =         {Early outcome-prediction with an automated {EEG}
                   background trend in hypothermia-treated newborns with
                   encephalopathy},
  volume =        {99},
  year          = {2026},
  doi =           {10.1038/s41390-025-04193-9},
  issn =          {1530-0447},
}

@article{Cornet2025,
  author =        {Cornet, Marie-Coralie and Numis, Adam L. and
                   Wusthoff, Courtney J. and Bernardo, Danilo and
                   Mietzsch, Ulrike and Thomas, Cameron and
                   Natarajan, Niranjana and Ahmad, Kaashif A. and
                   Scheffler, Aaron and Juul, Sandra E. and
                   Montazeri Moghadam, Saeed and Wu, Yvonne W. and
                   Glass, Hannah C.},
  journal =       {JAMA Network Open},
  month =         dec,
  number =        {12},
  pages =         {e2548321},
  publisher =     {American Medical Association (AMA)},
  title =         {Automated {EEG} Background Analysis and 2-Year
                   Outcomes in Neonatal Hypoxic-Ischemic Encephalopathy},
  volume =        {8},
  year =          {2025},
  doi =           {10.1001/jamanetworkopen.2025.48321},
  issn =          {2574-3805},
}

@article{Fanning2025,
  author =        {Fanning, Aisling and Pegorer, Elisabetta and
                   Pavel, Andreea M and Mathieson, Sean R and
                   Walsh, Brian H and Boylan, Geraldine B},
  journal =       {Seminars in Perinatology},
  month =         dec,
  number =        {8},
  pages =         {152163},
  publisher =     {Elsevier BV},
  title =         {Background {EEG} grading for the assessment of
                   neonatal encephalopathy in full term neonates},
  volume =        {49},
  year =          {2025},
  doi =           {10.1016/j.semperi.2025.152163},
  issn =          {0146-0005},
}

@article{Walsh2011,
  author        = {Walsh, B. H. and Murray, D. M. and Boylan, G. B.},
  journal =       {Clinical Neurophysiology},
  number =        {7},
  pages =         {1284--1294},
  publisher =     {Elsevier BV},
  title =         {The use of conventional {EEG} for the assessment of
                   hypoxic ischaemic encephalopathy in the newborn: A
                   review},
  volume =        {122},
  year =          {2011},
  doi =           {10.1016/j.clinph.2011.03.032},
  issn =          {1388-2457},
}

@article{Tsuchida2013,
  author =        {Tsuchida, Tammy N. and Wusthoff, Courtney J. and
                   Shellhaas, Renée A. and Abend, Nicholas S. and
                   Hahn, Cecil D. and Sullivan, Joseph E. and
                   Nguyen, Sylvie and Weinstein, Steven and
                   Scher, Mark S. and Riviello, James J. and
                   Clancy, Robert R.},
  journal =       {Journal of Clinical Neurophysiology},
  month =         apr,
  number =        {2},
  pages =         {161--173},
  publisher =     {Ovid Technologies (Wolters Kluwer Health)},
  title =         {American Clinical Neurophysiology Society
                   Standardized {EEG} Terminology and Categorization for
                   the Description of Continuous {EEG} Monitoring in
                   Neonates: Report of the {American Clinical
                   Neurophysiology Society Critical Care Monitoring
                   Committee}},
  volume =        {30},
  year =          {2013},
  doi =           {10.1097/wnp.0b013e3182872b24},
  issn =          {0736-0258},
}

@article{Mirzaei2017,
  author =        {Mirzaei, Parisa and Azemi, Ghasem},
  journal =       {Analog Integrated Circuits and Signal Processing},
  month =         may,
  number =        {2},
  pages =         {255--262},
  publisher =     {Springer Science and Business Media LLC},
  title =         {A new {NLEO} based technique for the detection of
                   burst–suppression patterns in multichannel neonatal
                   {EEG} signals},
  volume =        {92},
  year =          {2017},
  doi =           {10.1007/s10470-017-0989-0},
  issn =          {1573-1979},
}

@inproceedings{Raurale2019,
  author =        {Raurale, Sumit A. and Nalband, Saif and
                   Boylan, Geraldine B. and Lightbody, Gordon and
                   O’Toole, John M.},
  booktitle     = {Proc. IEEE Int. Conf. Engineering in Medicine and Biology Society (EMBC)},
  month =         jul,
  pages =         {4125--4128},
  publisher =     {IEEE},
  title =         {Suitability of an inter-burst detection method for
                   grading hypoxic-ischemic encephalopathy in newborn
                   {EEG}},
  year =          {2019},
  doi =           {10.1109/embc.2019.8857000},
}

@inproceedings{Raurale2020,
  author =        {Raurale, Sumit A. and Boylan, Geraldine B. and
                   Lightbody, Gordon and O’Toole, John M.},
  booktitle     = {Proc. IEEE Int. Conf. Engineering in Medicine and Biology Society (EMBC)},
  month =         jul,
  pages =         {5984--5987},
  publisher =     {IEEE},
  title =         {Identifying tracé alternant activity in neonatal
                   {EEG} using an inter-burst detection approach},
  year =          {2020},
  doi =           {10.1109/embc44109.2020.9176147},
}

@article{Lundy2023,
  author =        {Christopher Lundy and Geraldine B. Boylan and
                   Sean Mathieson and Jacopo Proietti and
                   John M. O{\textquotesingle}Toole},
  journal =       {Computers in Biology and Medicine},
  month =         oct,
  pages =         {107468},
  publisher =     {Elsevier {BV}},
  title =         {Quantitative analysis of high-frequency activity in
                   neonatal {EEG}},
  volume =        {165},
  year =          {2023},
  doi =           {10.1016/j.compbiomed.2023.107468},
}

@article{Hogan2025,
  author =        {Hogan, Robert and Mathieson, Sean R. and Luca, Aurel and
                   Ventura, Soraia and Griffin, Sean and
                   Boylan, Geraldine B. and O’Toole, John M.},
  journal =       {npj Digital Medicine},
  month =         jan,
  number =        {1},
  publisher =     {Springer Science and Business Media LLC},
  title =         {Scaling convolutional neural networks achieves expert
                   level seizure detection in neonatal {EEG}},
  volume =        {8},
  year =          {2025},
  doi =           {10.1038/s41746-024-01416-x},
  issn =          {2398-6352},
  pages         = {17},
}

@article{Korotchikova2016,
  author =        {Korotchikova, Irina and Stevenson, Nathan J. and
                   Livingstone, Vicki and Ryan, C. Anthony and
                   Boylan, Geraldine B.},
  journal =       {Clinical Neurophysiology},
  month =         apr,
  number =        {4},
  pages =         {2095--2101},
  publisher =     {Elsevier BV},
  title =         {Sleep–wake cycle of the healthy term newborn infant
                   in the immediate postnatal period},
  volume =        {127},
  year =          {2016},
  doi =           {10.1016/j.clinph.2015.12.015},
  issn =          {1388-2457},
}

@article{O’Toole2017,
  author =        {O’Toole, John M. and Boylan, Geraldine B. and
                   Lloyd, Rhodri O. and Goulding, Robert M. and
                   Vanhatalo, Sampsa and Stevenson, Nathan J.},
  journal =       {Medical Engineering and Physics},
  month =         apr,
  number =        {1},
  pages =         {42--50},
  publisher =     {IOP Publishing},
  title =         {Detecting bursts in the {EEG} of very and extremely
                   premature infants using a multi-feature approach},
  volume =        {45},
  year =          {2017},
  doi =           {10.1016/j.medengphy.2017.04.003},
  issn =          {1873-4030},
}

@article{Jennekens2011,
  author =        {Jennekens, Ward and Ruijs, Loes S and
                   Lommen, Charlotte M L and Niemarkt, Hendrik J and
                   Pasman, Jaco W and
                   van Kranen-Mastenbroek, Vivianne H J M and
                   Wijn, Pieter F F and van Pul, Carola and
                   Andriessen, Peter},
  journal =       {Physiological Measurement},
  month =         sep,
  number =        {10},
  pages =         {1623--1637},
  publisher =     {IOP Publishing},
  title =         {Automatic burst detection for the {EEG} of the
                   preterm infant},
  volume =        {32},
  year =          {2011},
  doi =           {10.1088/0967-3334/32/10/010},
  issn =          {1361-6579},
}

@inproceedings{O’Toole2014,
  author =        {O’Toole, John M. and Temko, Andriy and
                   Stevenson, Nathan},
  booktitle     = {Proc. IEEE Int. Conf. Engineering in Medicine and Biology Society (EMBC)},
  month =         aug,
  pages =         {3288--3291},
  publisher =     {IEEE},
  title =         {Assessing instantaneous energy in the {EEG}: A
                   non-negative, frequency-weighted energy operator},
  year =          {2014},
  doi =           {10.1109/embc.2014.6944325},
}

@article{Koolen2014,
  author =        {Koolen, Ninah and Jansen, Katrien and Vervisch, Jan and
                   Matic, Vladimir and De Vos, Maarten and
                   Naulaers, Gunnar and Van Huffel, Sabine},
  journal =       {Clinical Neurophysiology},
  month =         oct,
  number =        {10},
  pages =         {1985--1994},
  publisher =     {Elsevier BV},
  title =         {Line length as a robust method to detect
                   high-activity events: Automated burst detection in
                   premature {EEG} recordings},
  volume =        {125},
  year =          {2014},
  doi =           {10.1016/j.clinph.2014.02.015},
  issn =          {1388-2457},
}

@article{Matic2016,
  author =        {Matic, Vladimir and Cherian, Perumpillichira J. and
                   Jansen, Katrien and Koolen, Ninah and
                   Naulaers, Gunnar and Swarte, Renate M. and
                   Govaert, Paul and Van Huffel, Sabine and
                   De Vos, Maarten},
  journal =       {IEEE Transactions on Biomedical Engineering},
  month =         may,
  number =        {5},
  pages =         {973--983},
  publisher =     {Institute of Electrical and Electronics Engineers
                   (IEEE)},
  title =         {Improving Reliability of Monitoring Background {EEG}
                   Dynamics in Asphyxiated Infants},
  volume =        {63},
  year =          {2016},
  doi =           {10.1109/tbme.2015.2477946},
  issn =          {1558-2531},
}

@article{Palmu2010,
  author =        {Palmu, Kirsi and Wikström, Sverre and
                   Hippeläinen, Eero and Boylan, Geraldine and
                   Hellström-Westas, Lena and Vanhatalo, Sampsa},
  journal =       {Clinical Neurophysiology},
  month =         jul,
  number =        {7},
  pages =         {1015--1022},
  publisher =     {Elsevier BV},
  title =         {Detection of ‘{EEG} bursts’ in the early preterm
                   {EEG}: Visual vs. automated detection},
  volume =        {121},
  year =          {2010},
  doi =           {10.1016/j.clinph.2010.02.010},
  issn =          {1388-2457},
}

@article{Palmu2010a,
  author =        {Palmu, Kirsi and Stevenson, Nathan and
                   Wikström, Sverre and Hellström-Westas, Lena and
                   Vanhatalo, Sampsa and Palva, J Matias},
  journal =       {Physiological Measurement},
  month =         oct,
  number =        {11},
  pages =         {N85--N93},
  publisher =     {IOP Publishing},
  title =         {Optimization of an {NLEO}-based algorithm for
                   automated detection of spontaneous activity
                   transients in early preterm {EEG}},
  volume =        {31},
  year =          {2010},
  doi =           {10.1088/0967-3334/31/11/n02},
  issn =          {1361-6579},
}

@inproceedings{Liu2022,
  author =        {Liu, Zhuang and Mao, Hanzi and Wu, Chao-Yuan and
                   Feichtenhofer, Christoph and Darrell, Trevor and
                   Xie, Saining},
  booktitle     = {Proc. IEEE/CVF Conf. Computer Vision and Pattern Recognition (CVPR)},
  month =         jun,
  pages =         {11976--11986},
  title =         {A {ConvNet} for the 2020s},
  year =          {2022},
}

@article{DeVries2017,
  author =        {DeVries, Terrance and Taylor, Graham W.},
  journal =       {arXiv:1708.04552},
  publisher =     {arXiv},
  title =         {Improved Regularization of Convolutional Neural
                   Networks with Cutout},
  year =          {2017},
  doi =           {10.48550/ARXIV.1708.04552},
}

@article{Kljajic2025,
  author =        {Kljajic, Jovana and O'Toole, John M. and
                   Hogan, Robert and Skoric, Tamara},
  journal =       {Medical Engineering and Physics},
  title =         {Honest and Reliable Evaluation and Expert Equivalence
                   Testing of Automated Neonatal Seizure Detection},
  year =          {2026},
  publisher =     {{IOP}},
  doi =           {10.1088/1873-4030/aea852},
  note          = {In press. DOI: 10.1088/1873-4030/aea852},
}

@article{Tapani2022,
  author =        {Karoliina T. Tapani and Päivi Nevalainen and
                   Sampsa Vanhatalo and Nathan J. Stevenson},
  journal =       {Computers in Biology and Medicine},
  month =         jun,
  pages =         {105399},
  publisher =     {Elsevier {BV}},
  title =         {Validating an {SVM}-based neonatal seizure detection
                   algorithm for generalizability, non-inferiority and
                   clinical efficacy},
  volume =        {145},
  year =          {2022},
  doi =           {10.1016/j.compbiomed.2022.105399},
}

@article{OToole2023,
  author =        {John M. O'Toole and Sean R. Mathieson and
                   Sumit A. Raurale and Fabio Magarelli and
                   William P. Marnane and Gordon Lightbody and
                   Geraldine B. Boylan},
  journal =       {Scientific Data},
  month =         mar,
  number =        {1},
  publisher =     {Springer Science and Business Media {LLC}},
  title =         {Neonatal {EEG} graded for severity of background
                   abnormalities in hypoxic-ischaemic encephalopathy},
  volume =        {10},
  year =          {2023},
  doi =           {10.1038/s41597-023-02002-8},
  pages         = {129},
}

@article{Ziv1978,
  author =        {Ziv, J. and Lempel, A.},
  journal =       {IEEE Transactions on Information Theory},
  number =        {5},
  pages =         {530--536},
  publisher =     {Institute of Electrical and Electronics Engineers
                   (IEEE)},
  title =         {Compression of individual sequences via variable-rate
                   coding},
  volume =        {24},
  year =          {1978},
  doi =           {10.1109/tit.1978.1055934},
  issn =          {1557-9654},
}

@article{Pavel2022,
  author =        {Andreea M. Pavel and John M. O{\textquotesingle}Toole and
                   Jacopo Proietti and Vicki Livingstone and
                   Subhabrata Mitra and William P. Marnane and
                   Mikael Finder and Eugene M. Dempsey and
                   Deirdre M. Murray and Geraldine B. Boylan},
  journal =       {Epilepsia},
  month         = feb,
  number =        {2},
  pages =         {456--468},
  publisher =     {Wiley},
  title =         {Machine learning for the early prediction of infants
                   with electrographic seizures in neonatal
                   hypoxic-ischemic encephalopathy},
  volume =        {64},
  year          = {2023},
  doi =           {10.1111/epi.17468},
}

@article{Mader2024,
  author =        {Mader, Johannes and Hartmann, Manfred and
                   Dressler, Anastasia and Oberdorfer, Lisa and
                   Rona, Zsofia and Glatter, Sarah and
                   Czaba-Hnizdo, Christine and Herta, Johannes and
                   Kluge, Tilmann and Werther, Tobias and
                   Berger, Angelika and Koren, Johannes and
                   Klebermass-Schrehof, Katrin and Giordano, Vito},
  journal =       {Physiological Measurement},
  month =         sep,
  number =        {9},
  pages =         {095017},
  publisher =     {IOP Publishing},
  title =         {Adaptive threshold algorithm for detecting
                   {EEG}-interburst intervals in extremely preterm
                   neonates},
  volume =        {45},
  year =          {2024},
  doi =           {10.1088/1361-6579/ad7c05},
  issn =          {1361-6579},
}

@article{OToole2017,
  author =        {O'Toole, John M. and Boylan, Geraldine B.},
  journal =       {arXiv:1704.05694},
  month =         apr,
  publisher =     {arXiv},
  title =         {{NEURAL}: quantitative features for newborn {EEG} using
                   {Matlab}},
  year =          {2017},
  doi =           {10.48550/ARXIV.1704.05694},
}

@article{Stevenson2012,
  author =        {Stevenson, N. J. and Korotchikova, I. and Temko, A. and
                   Lightbody, G. and Marnane, W. P. and Boylan, G. B.},
  journal =       {Annals of Biomedical Engineering},
  month         = apr,
  number =        {4},
  pages =         {775--785},
  publisher =     {Springer Science and Business Media LLC},
  title =         {An Automated System for Grading {EEG} Abnormality in
                   Term Neonates with Hypoxic-Ischaemic Encephalopathy},
  volume =        {41},
  year          = {2013},
  doi =           {10.1007/s10439-012-0710-5},
  issn =          {1573-9686},
}

@article{Matic2014,
  author =        {Matic, Vladimir and Cherian, Perumpillichira J and
                   Koolen, Ninah and Naulaers, Gunnar and
                   Swarte, Renate M and Govaert, Paul and
                   Van Huffel, Sabine and De Vos, Maarten},
  journal =       {Journal of Neural Engineering},
  month =         oct,
  number =        {6},
  pages =         {066007},
  publisher =     {IOP Publishing},
  title =         {Holistic approach for automated background {EEG}
                   assessment in asphyxiated full-term infants},
  volume =        {11},
  year =          {2014},
  doi =           {10.1088/1741-2560/11/6/066007},
  issn =          {1741-2552},
}

@article{Raurale2021,
  author =        {Sumit A Raurale and Geraldine B Boylan and
                   Sean R Mathieson and William P Marnane and
                   Gordon Lightbody and John M O'Toole},
  journal =       {Journal of Neural Engineering},
  month =         mar,
  number =        {4},
  pages =         {046007},
  publisher =     {{IOP} Publishing},
  title =         {Grading hypoxic-ischemic encephalopathy in neonatal
                   {EEG} with convolutional neural networks and
                   quadratic time{\textendash}frequency distributions},
  volume =        {18},
  year =          {2021},
  doi =           {10.1088/1741-2552/abe8ae},
}

@article{Dilena2024,
  author =        {Dilena, Robertino and Cilio, Maria Roberta},
  journal =       {Pediatric Research},
  month =         aug,
  number =        {4},
  pages =         {841--843},
  publisher =     {Springer Science and Business Media LLC},
  title =         {Free access via computational cloud to deep
                   learning-based {EEG} assessment in neonatal
                   hypoxic-ischemic encephalopathy: revolutionary
                   opportunities to overcome health disparities},
  volume =        {96},
  year =          {2024},
  doi =           {10.1038/s41390-024-03427-6},
  issn =          {1530-0447},
}

@article{Bernardo2024,
  author =        {Bernardo, Danilo and Kim, Jonathan and
                   Cornet, Marie‐Coralie and Numis, Adam L. and
                   Scheffler, Aaron and Rao, Vikram R. and
                   Amorim, Edilberto and Glass, Hannah C.},
  journal =       {Epilepsia},
  month         = jan,
  number =        {1},
  pages =         {89--103},
  publisher =     {Wiley},
  title =         {Machine learning for forecasting initial seizure
                   onset in neonatal hypoxic–ischemic encephalopathy},
  volume =        {66},
  year          = {2025},
  doi =           {10.1111/epi.18163},
  issn =          {1528-1167},
}

@inproceedings{Agarwal,
  author =        {Agarwal, R. and Gotman, J.},
  booktitle     = {Proc. IEEE Int. Symp. Circuits and Systems (ISCAS)},
  pages =         {199--202},
  publisher =     {IEEE},
  title =         {Adaptive segmentation of electroencephalographic data
                   using a nonlinear energy operator},
  volume =        {4},
  year =          {1999},
  doi =           {10.1109/iscas.1999.779976},
}

@inproceedings{Matic2012,
  author =        {Matic, Vladimir and Cherian, P. J. and Jansen, K. and
                   Koolen, N. and Naulaers, G. and Swarte, R. M. and
                   Govaert, P. and Visser, G. H. and Van Huffel, S. and
                   De Vos, M.},
  booktitle     = {Proc. IEEE Int. Conf. Engineering in Medicine and Biology Society (EMBC)},
  month         = aug,
  pages         = {17--20},
  publisher =     {IEEE},
  title =         {Automated {EEG} inter-burst interval detection in
                   neonates with mild to moderate postasphyxial
                   encephalopathy},
  year =          {2012},
  doi =           {10.1109/embc.2012.6345860},
}

@article{Saerkelae2002,
  author =        {Särkelä, Mika and Mustola, Seppo and
                   Seppänen, Tapio and Koskinen, Miika and Lepola, Pasi and
                   Suominen, Kalervo and Juvonen, Tatu and
                   Tolvanen-Laakso, Heli and Jäntti, Ville},
  journal =       {Journal of Clinical Monitoring and Computing},
  month =         feb,
  number =        {2},
  pages =         {125--134},
  publisher =     {Springer Science and Business Media LLC},
  title =         {Automatic Analysis and Monitoring of Burst
                   Suppression in Anesthesia},
  volume =        {17},
  year =          {2002},
  doi =           {10.1023/a:1016393904439},
  issn =          {1573-2614},
}

\clearpage
\appendix
\makeatletter
\renewcommand{\fps@table}{H}
\renewcommand{\fps@figure}{H}
\makeatother
\renewcommand{\thetable}{A\arabic{table}}
\setcounter{table}{0}
\renewcommand{\thefigure}{A\arabic{figure}}
\setcounter{figure}{0}
\FloatBarrier

\section*{Appendix}
\renewcommand{\thesubsection}{A.\arabic{subsection}}
\renewcommand{\theequation}{A.\arabic{equation}}
\setcounter{equation}{0}

\subsection{Quantitative Features for IBI Detection}
\label{sec-app:quant-feat-ibi}

IBIs are defined by amplitude.
Not surprising, therefore, that many methods for detecting these intervals are amplitude-based measures.
Variations in how to quantify amplitude, either through peak amplitude or average power measures \cite{Matic2016,Mader2024,Jennekens2011}, have been applied to both preterm and term neonatal EEG.
A direct approach is to simply measure the ratio of signal outside the \qty{25}{\micro\volt} peak-to-peak range over a 1 s window
\begin{equation}
  \label{eq:5}
  P(n) = \frac{1}{f_s} \sum_{m=1}^{f_s} \mathbb{I}[|x(n-m)| > V_{\text{th}}]
\end{equation}
where $f_s$ is the sampling frequency of signal $x(n)$ with $n$ as discrete time, $\mathbb{I}[\cdot]$ is the indicator function, and $V_{\text{th}}=12.5$ \qty{}{\micro\volt}.
Amplitude can also be measured through the envelope of the signal, defined as
\begin{equation}
  \label{eq:3}
  E(n) = \left| x(n) + j\mathcal{H}[x(n)]\right|
\end{equation}
for Hilbert transform $\mathcal{H}[\cdot]$.

Other methods incorporate frequency into the amplitude measure.
These frequency-weighted energy measures align with the physics definition of energy, as higher-frequency signals require more energy to generate than lower-frequency signals of the same amplitude.
The nonlinear energy operator, proposed in 1999 for adult EEG \cite{Agarwal}, was the first example of a frequency-weighted energy measure applied to EEG.
Defined as
\begin{equation*}
\Theta[x(n)] = x(n-1)x(n-2) - x(n)x(n-3),
\end{equation*}
this measure can oscillate widely and can produce negative values, thus violating a basic property of an energy measure.
Post-processing, such as smoothing the absolute value, is applied to overcome these limitations.
This operator has remained a popular choice for the detection of bursts (or IBIs) in preterm and term EEG \cite{Palmu2010,Palmu2010a,Matic2012,Mirzaei2017}.
To the best of our knowledge, it is also the only IBI detector used in a medical device.
It is implemented in the EEG NicoletOne ICU Monitor (Natus, USA) as a burst-suppression detector
\cite{Saerkelae2002,Palmu2010} and a modified version \cite{Palmu2010} is implemented in the ANT
Neuro n{\"e}o monitor (eemagine Medical Imaging Solutions GmbH, Berlin, Germany) for detection of
IBIs and calculation of burst suppression.

Line-length, another frequency-weighted energy measure developed for neonatal EEG, is defined as the absolute value of the difference summed across a \qty{1}{\second} window
\begin{equation}
  \label{eq:4}
  L(n) = \sum_{m=1}^{f_s} |x(n-m+1) - x(n-m)|.
\end{equation}
The original definition included a normalisation factor \cite{Koolen2014}; however, we found that this significantly degraded performance for LVI detection and therefore we have not included it here in this definition.

The last feature we present here is a variant of the nonlinear energy operator, known as the envelope derivative operator (EDO) \cite{O’Toole2014}.
It fixes some of the issues with the nonlinear method without the need for pre- (band-pass filtering) and post-processing \cite{Palmu2010a} and retains desirable properties such as non-negativity as an instantaneous energy measure.
It is defined as
\begin{eqnarray}
  \label{eq:2}
    \Gamma[x(n)] &= & \frac{1}{4}\left[x(n+1)^2 + x(n-1)^2 + h(n+1)^2 + h(n-1)^2\right] - \nonumber\\
     && \frac{1}{2}\left[x(n+1)x(n-1) + h(n+1)h(n-1)\right]
\end{eqnarray}
where $h(n)=\mathcal{H}[x(n)]$, the Hilbert transform.

Implementation of these measures as a detector, which are continuous unbounded values, requires a threshold to convert to binary masks.
Although adaptive thresholds are often proposed as a better solution than a static threshold \cite{Koolen2014,Mader2024}, this requires many assumptions about the EEG characteristics of the epoch being processed.
Unsurprisingly, we found poor generalisation performance with adaptive thresholding in our
preliminary testing.

\subsection{Results}

\begin{table}
  \centering
  \scriptsize
  \caption{Demographic information on the EEG data sets used for the application of the
    LVI detectors.}
  \label{tab:evaluation_set}
  \begin{threeparttable}
    \begin{tabular}{lr}
      \toprule
                                              & application set             \\
                                              & ($n=391$)                     \\ 
      \midrule
      Gestational Age (weeks)                 & 40+0 (39+1 -- 41+0) [$n=386$] \\
      Birth Weight (kg)                       & 3.5 (3.14 -- 3.79) [$n=151$]  \\
      Sex, Female                             & 138 (43.4\%) [$n=318$]        \\
      HIE                                     & 222 (56.8\%)                \\
      HIE: Mild / Moderate / Severe / Unknown & 96 / 94 / 31 / 1            \\
      Normal cohort                           & 48 (12.3\%)                 \\
      \midrule
      Total EEG duration (hours)              & 18,825                      \\
      \bottomrule
    \end{tabular}
    \begin{tablenotes}
      \scriptsize
      \item HIE: hypoxic ischaemic encephalopathy
    \end{tablenotes}
  \end{threeparttable}
\end{table}

\begin{table}
  \centering
  \scriptsize
  \caption{Summary of LVI annotations for the development and validation datasets.}
  \label{tab:lvi_annotation_intervals}
  \begin{threeparttable}
    \begin{tabular}{lrrrr}
      \toprule
      Dataset     & Reviewer & Number of  & Prevalence & Duration (seconds) \\
                  &          & Events     &            & median (IQR) \\
      \midrule
      Development & R1       & 36,580  & 0.202      & 5.48 (3.55--9.31)     \\
      Validation  & R2       & 5,845   & 0.306      & 4.59 (2.94--9.19)     \\
      Validation  & R3       & 8,737   & 0.299      & 3.20 (2.36--6.06)     \\
      \bottomrule
    \end{tabular}
    \begin{tablenotes}
      \scriptsize
    \item IQR: interquartile range
    \end{tablenotes}
  \end{threeparttable}
\end{table}

\begin{figure}
  \centering
  \includegraphics[width=1.00\textwidth]{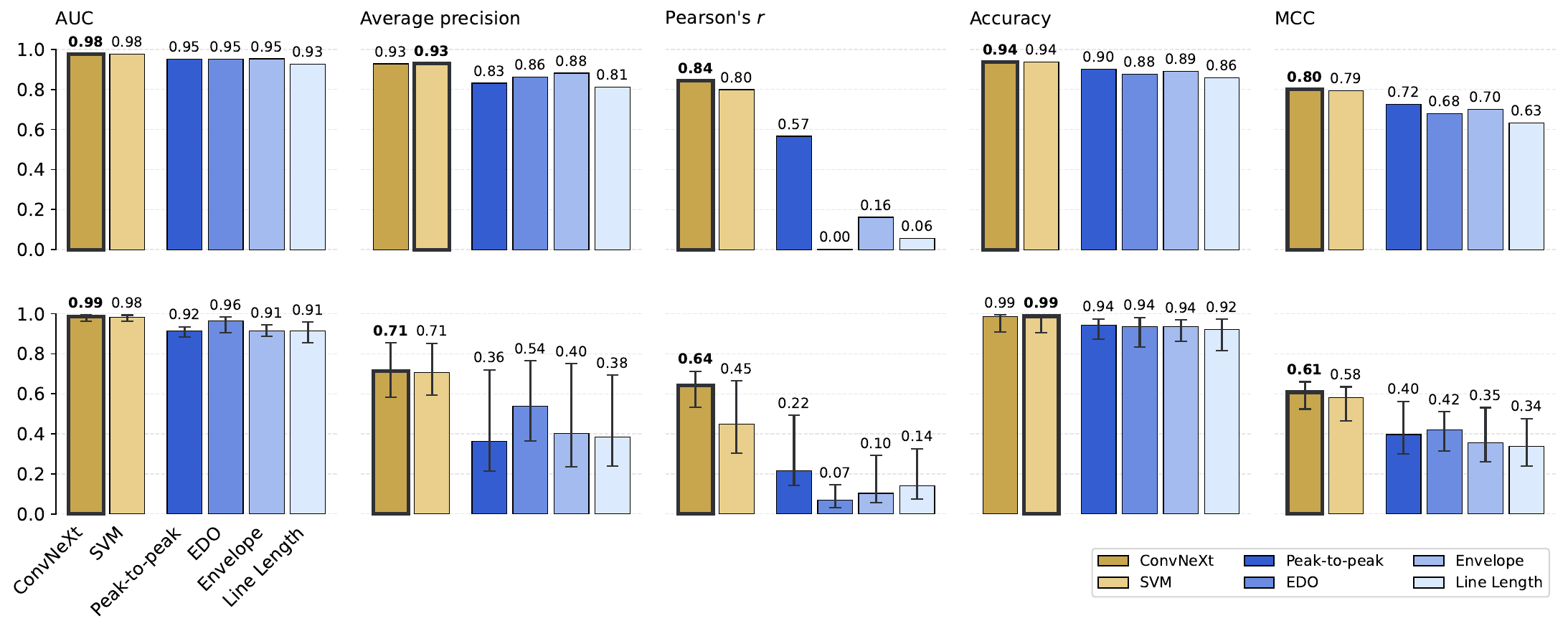}
  \caption{Cross-validation testing results for the LVI detectors across the 10 cross-validation folds.
Performance metrics are calculated from the predictions concatenated across all EEG records in the top row and on a per-neonate basis in the bottom row.
For the per-neonate results, metrics are summarised by median with error bars representing the interquartile range.}
  \label{fig:ibi_xv_results}
\end{figure}

\begin{table}[H]
  \centering
  \scriptsize
  \caption{Held-out validation performance of different LVI detectors.
Data are concatenated across all EEG files.
Metrics calculated using consensus annotations from two human experts.}
    \label{tab:heldout_concat_results}
    \begin{threeparttable}    
    \begin{tabular}{lrrrrrrrr}
      \toprule
              & AUC         & AP          & $r$         & accuracy   & MCC         & $k$         & sens./spec.       & PPV/NPV           \\
              &             &             &             & (\%)       &             &             & (\%)              & (\%)              \\
      \midrule
 ConvNeXt     & {\bf 0.996} & {\bf 0.943} & {\bf 0.992} & {\bf 96.9} & {\bf 0.926} & {\bf 0.926} & 95.9 / 97.4       & 93.7 / 98.3       \\
 SVM          & {\bf 0.996} & 0.895       & 0.990       & 96.8       & 0.923       & 0.923       & 94.4 / {\bf 97.8} & {\bf 94.6} / 97.7 \\
 Peak-to-peak & 0.983       & 0.673       & 0.943       & 93.0       & 0.845       & 0.838       & 97.5 / 91.1       & 81.8 / 98.9       \\
 EDO          & 0.988       & 0.014       & 0.971       & 91.1       & 0.813       & 0.800       & {\bf 97.9} / 88.4 & 77.5 / {\bf 99.0} \\
 Envelope     & 0.987       & 0.176       & 0.975       & 91.7       & 0.819       & 0.810       & 96.9 / 89.5       & 79.0 / 98.6       \\
 Line Length  & 0.984       & 0.145       & 0.965       & 89.3       & 0.780       & 0.763       & 97.8 / 85.8       & 73.7 / {\bf 99.0} \\
      \bottomrule
    \end{tabular}
    \begin{tablenotes}
      \scriptsize
    \item Key: AUC, area under the receiver operating characteristic curve (AUC); AP, average
      precision; $r$, Pearson's correlation coefficient; MCC, Matthews correlation coefficient;
      $k$, Cohen's kappa; sens., sensitivity; spec., specificity; PPV, positive predictive value;
      NPV, negative predictive value; SVM, support vector machine; EDO, energy derivative operator
      \end{tablenotes}
  \end{threeparttable}

\end{table}

\begin{table}[H]
  \centering
  \scriptsize
    \caption{Held-out validation performance of different LVI detectors. Data are summarised per
      neonate as median (interquartile range).
      Metrics calculated using consensus annotations from two human experts.}
    \label{tab:heldout_perbaby_results}    
    \begin{threeparttable}    
    \begin{tabular}{lrrrr}
      \toprule
              & AUC                    & $r$                    & accuracy (\%)          & MCC                    \\
      \midrule
 ConvNeXt     & {\bf 0.994}            & {\bf 0.788}            & {\bf 98.7}            & {\bf 0.736}            \\
              & {\bf (0.987 to 0.998)} & {\bf (0.652 to 0.867)} & {\bf (95.7 to 99.6)} & {\bf (0.629 to 0.849)} \\
 SVM          & 0.993                  & 0.559                  & 98.6                  & 0.703                  \\
              & (0.985 to 0.996)       & (0.335 to 0.750)       & (95.2 to 99.6)       & (0.556 to 0.812)       \\
 Peak-to-peak & 0.955                  & 0.319                  & 93.8                  & 0.470                  \\
              & (0.905 to 0.975)       & (0.154 to 0.503)       & (90.1 to 97.9)       & (0.359 to 0.676)       \\
 EDO          & 0.964                  & 0.172                  & 92.7                  & 0.381                  \\
              & (0.931 to 0.990)       & (0.048 to 0.359)       & (86.6 to 98.2)       & (0.285 to 0.557)       \\
 Envelope     & 0.959                  & 0.192                  & 91.8                  & 0.421                  \\
              & (0.914 to 0.981)       & (0.061 to 0.340)       & (88.4 to 97.2)       & (0.289 to 0.614)       \\
 Line Length  & 0.947                  & 0.228                  & 90.4                  & 0.342                  \\
              & (0.921 to 0.979)       & (0.084 to 0.462)       & (80.3 to 97.0)       & (0.245 to 0.543)       \\
      \bottomrule
    \end{tabular}
    \begin{tablenotes}
      \scriptsize
    \item Key: AUC, area under the receiver operating characteristic curve (AUC); $r$, Pearson's
      correlation coefficient; MCC, Matthews correlation coefficient; SVM, support vector machine;
      EDO, energy derivative operator
    \end{tablenotes}
  \end{threeparttable}
\end{table}

\begin{table}
  \centering
  \scriptsize
  \caption{Expert-level equivalence testing of low-voltage interval detection on the held-out validation set.
Mean changes in Cohen's kappa ($\Delta\kappa$) and 95\% confidence intervals (CI) quantify the change in inter-rater agreement when one human annotation is replaced by detector predictions, relative to agreement between the two human experts.
Positive values of $\Delta\kappa$ indicate an increase in agreement between the detector and each human annotation.
Only the ConvNeXt and the SVM detectors reach expert-level agreement.
Distributions are generated using a bootstrapping process with 10k sampling iterations.}
\label{sec:expert_testing}
  \begin{threeparttable}
    \begin{tabular}{lrrr}
      \toprule
                   & $\Delta \kappa$ & 95\% CI              & $p$-value \\
      \midrule
      ConvNeXt     & {\bf 0.024}     & {\bf 0.011 to 0.040} & $<$0.001  \\
      SVM          & 0.020           & 0.008 to 0.034       & $<$0.001  \\
      Peak-to-peak & -0.049          & -0.073 to -0.025     & 0.774     \\
      EDO          & -0.081          & -0.104 to -0.058     & 0.999     \\
      Envelope     & -0.074          & -0.100 to -0.049     & 0.991     \\
      Line Length  & -0.114          & -0.140 to -0.087     & $>$0.999  \\
      \bottomrule
    \end{tabular}
    \begin{tablenotes}
      \scriptsize
      \item Key: SVM, support vector machine; EDO, energy derivative operator.
    \end{tablenotes}
  \end{threeparttable}
\end{table}

\begin{table}[H]
  \centering
  \scriptsize
  \caption{ Difference in LVI features calculated on the held-out validation set comparing the features generated from the ConvNeXt prediction mask to the average of the two human experts.
Mean differences and 95\% confidence intervals (CI) were generated from bootstrapped sampling across neonates.
A one-sided non-inferiority test determines whether ML model error was greater than or equal to human inter-rater error; small $p$-values indicate that model error was consistently lower than, or comparable to, expert inter-rater variability.
}
  \label{tab:lvi_feature_differences}
  \begin{threeparttable}
    \begin{tabular}{lrrr}
      \toprule
                           & Mean difference & 95\% CI              & $p$-value \\
      \midrule
      Maximum duration (s) &   0.100  & -0.686 to 0.856       & 0.583     \\      
      Median duration (s)  &   0.172  & -0.188 to 0.594       & 0.777     \\
      Percentage (\%)      &  -1.291  & -2.818 to -0.245      & 0.023     \\      
      Number of events     & -24.489  & -34.558 to -11.589    & $<$0.001  \\
      Complexity           &  -0.036  & -0.068 to -0.019      & 0.001     \\
      Periodicity          &  -0.001  & -0.002 to -0.000      & 0.013     \\
      \bottomrule
    \end{tabular}
  \end{threeparttable}
\end{table}

\begin{table}[H]
  \centering
  \scriptsize
  \caption{Measures of separation for the six LVI features across (a) the four EEG background grades and (b) the three clinical HIE grades and the normal cohort.
The effect-size measure $\varepsilon^2$ quantifies the difference across the four grades and Spearman's correlation coefficient ($\rho$) quantifies the monotonic relationship across grades.
}
  \label{tab:lvi_app_eeg_grades}
  \begin{threeparttable}
    \begin{tabular}{lrrrr}
      \toprule
      & \multicolumn{2}{c}{EEG background grades} & \multicolumn{2}{c}{HIE grades \& normal cohort} \\
      \cmidrule(lr){2-3} \cmidrule(lr){4-5}
      feature              & effect size       & correlation & effect size       & correlation \\
                           & ($\varepsilon^2$) & ($\rho$)    & ($\varepsilon^2$) & ($\rho$)    \\
      \midrule                                                                   
      Maximum duration (s) & 0.521             & 0.714       & 0.269             & 0.311       \\
      Median duration (s)  & 0.676             & 0.821       & 0.248             & 0.349       \\
      Percent (\%)         & 0.643             & 0.801       & 0.297             & 0.384       \\
      Number               & 0.523             & 0.630       & 0.164             & 0.250       \\
      Complexity           & 0.538             & 0.667       & 0.143             & 0.262       \\
      Periodicity          & 0.405             & 0.626       & 0.102             & 0.223       \\
      \bottomrule
    \end{tabular}
    \begin{tablenotes}
      \item All measures are statistically significant with $p<0.001$.
    \end{tablenotes}    
  \end{threeparttable}
\end{table}

\begin{table}[H]
  \centering
  \scriptsize
  \caption{Measures of association for the six LVI features for a) the seizure prediction task (203 neonates) and b) association with seizure burden (121 neonates).
The area under the receiver operating characteristic curve (AUC) measures detection performance, the effect-size measure ($\varepsilon^2$) quantifies the difference between groups, and Spearman's correlation coefficient ($\rho$) quantifies the monotonic relationship between the seizure-neonate groups and seizure burden.
}\label{tab:lvi_app_sz}
  \begin{threeparttable}
    \begin{tabular}{lllll}
      \toprule
      & \multicolumn{2}{c}{Seizure prediction} & \multicolumn{2}{c}{Association with seizure burden} \\
      \cmidrule(lr){2-3} \cmidrule(lr){4-5}
      feature              & AUC       & correlation & effect size       & correlation \\
                           &  & ($\rho$)    & ($\varepsilon^2$) & ($\rho$)    \\
      \midrule                                                                   
      Maximum duration (s) & $0.850^{***}$ & $0.447^{***}$ & $0.116^{**}$ & $0.258^{**}$     \\
      Median duration (s)  & $0.841^{***}$ & $0.437^{***}$ & $0.136^{**}$ & $0.294^{**}$       \\
      Percent (\%)         & $0.872^{***}$ & $0.475^{***}$ & $0.136^{**}$ & $0.255^{**}$       \\
      Number               & $0.816^{***}$ & $0.404^{***}$ & $0.120^{**}$ & $0.163$  \\
      Complexity           & $0.816^{***}$ & $0.405^{***}$ & $0.098^{**}$ & $0.131$  \\
      Periodicity          & $0.830^{***}$ & $0.423^{***}$ & $0.138^{**}$ & $0.398^{***}$ \\
      \bottomrule
    \end{tabular}
    \begin{tablenotes}
    \item Significance levels are denoted as $^{*}$ $p<0.05$, $^{**}$ $p<0.01$, $^{***}$ $p<0.001$.
    \end{tablenotes}    
  \end{threeparttable}
\end{table}

\begin{figure}
  \centering
  \includegraphics[width=1.00\textwidth]{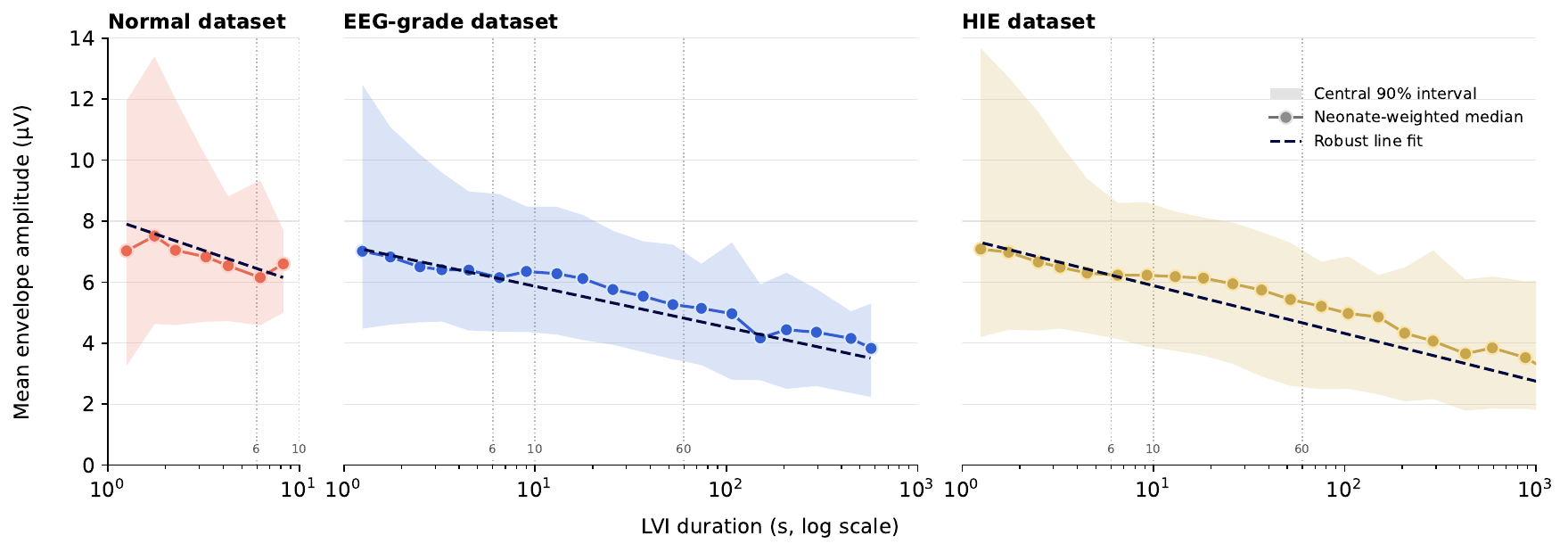}
  \footnotesize
  \caption{Association between the duration of LVIs and EEG amplitude.
Across all three datasets, the trend indicates decreasing amplitude with increasing duration.
Amplitude for each LVI was measured as the mean value of the EEG envelope.
LVI events were grouped into half-octave bins on the logarithmic duration scale.
For each bin, amplitude--duration pairs were weighted inversely by the number of events contributed by each neonate, ensuring that every neonate contributed equal total weight within that bin.
The neonate-weighted median amplitude and median duration for each bin were plotted with an amplitude interval bounded by the weighted 5th and 95th percentiles.
Bins with fewer than five neonates were excluded from the summary.
A line was fitted to the un-binned data pairs using a Huber loss function to limit the influence of outliers.
The model fitted amplitude as a linear function of a $\log_{10}$ transformed duration.
Amplitude--duration pairs were weighted inversely by each neonate's total number of LVIs, giving every neonate equal total influence on the fitted line.
}
  \label{fig:ibi_duration_amplitude}
\end{figure}

\begin{figure}
  \centering
  \includegraphics[width=1.00\textwidth]{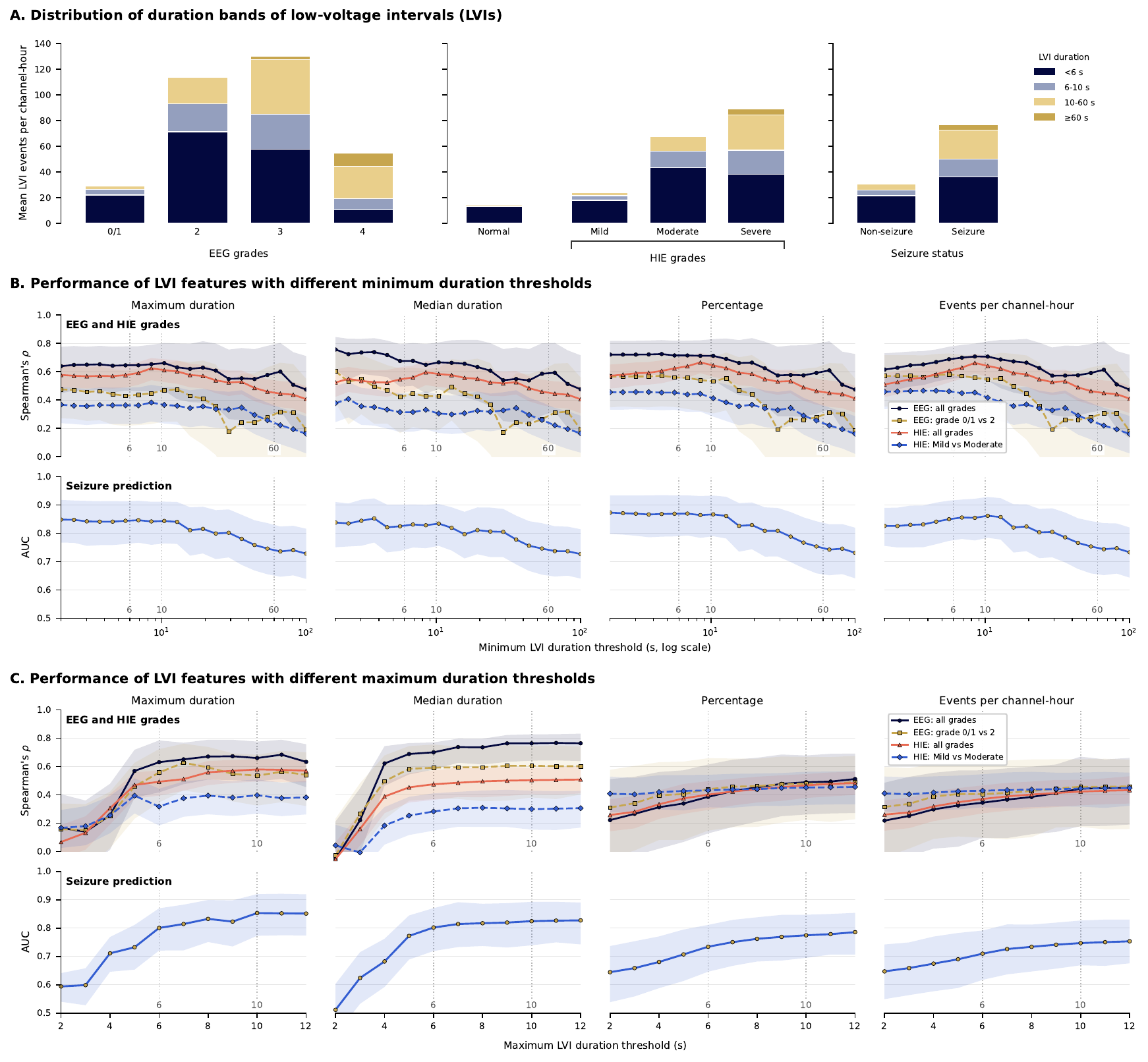}
  \footnotesize
  \caption{Analysis of LVI duration.
A: Distribution of the different categories of LVI duration across three datasets.
The number of LVIs within a duration category is calculated per neonate, normalised to the total available EEG channel-hours.
The stacked bars display the mean per-neonate event rate for each duration category within each grade or class.
B: Performance of LVI features when setting a minimum-duration threshold on included LVI events.
LVIs shorter than this threshold were removed, therefore testing the importance of, or robustness to, including shorter-duration LVIs.
Three applications are included, namely the EEG grades of background activity, clinical HIE grades, and a seizure prediction task.
In addition to the ordinal analysis across all grades for EEG and HIE, a secondary binary comparison is included to distinguish specific severity sub-groups: grade 0/1 versus 2 for EEG grades and mild versus moderate HIE.
}
\label{fig:ibi_duration_analysis}
\end{figure}



\end{document}